\documentclass{article}

\usepackage{amsmath,amsfonts,bm,dsfont}

\def\boml{\textsc{Boml}}
\def\bomla{\textsc{BomLA}}
\def\bomvi{\textsc{BomVI}}
\DeclareMathOperator{\vectorise}{vec}

\newcommand{\dtset}{\mathcal{D}}
\newcommand{\task}{\mathcal{T}}
\def\Appref#1{Appendix~\ref{#1}}

\def\Tabref#1{Table~\ref{#1}}
\def\Twotabref#1#2{Tables~\ref{#1} and \ref{#2}}

\newcommand{\newterm}[1]{{\bf #1}}

\def\Figref#1{Figure~\ref{#1}}

\def\Secref#1{Section~\ref{#1}}

\def\eqref#1{equation~\ref{#1}}
\def\Eqref#1{Eq.~(\ref{#1})}                                 

\def\Algref#1{Algorithm~\ref{#1}}

\def\Twoalgref#1#2{Algorithms \ref{#1} and \ref{#2}}

\def\1{\bm{1}}

\DeclareMathAlphabet{\mathsfit}{\encodingdefault}{\sfdefault}{m}{sl}
\SetMathAlphabet{\mathsfit}{bold}{\encodingdefault}{\sfdefault}{bx}{n}

\newcommand{\E}{\mathbb{E}}

\newcommand{\KL}{D_{\mathrm{KL}}}

\DeclareMathOperator*{\argmax}{arg\,max}
\DeclareMathOperator*{\argmin}{arg\,min}

\usepackage{bm}
\usepackage{microtype}
\usepackage{graphicx}
\usepackage{subfigure}
\usepackage{booktabs} 
\usepackage{gensymb} 
\usepackage{spverbatim}
\usepackage{algorithm}
\usepackage{algorithmicx}
\usepackage{algpseudocode}
\usepackage{hyperref}
\usepackage{rotating}
\usepackage{xcolor}
\usepackage{url}
\usepackage{stfloats}

\usepackage{hyperref}

\usepackage[accepted]{icml2021}

\icmltitlerunning{Addressing Catastrophic Forgetting in Few-Shot Problems}

\begin{document}

\twocolumn[
\icmltitle{Addressing Catastrophic Forgetting in Few-Shot Problems}



\icmlsetsymbol{equal}{*}

\begin{icmlauthorlist}
\icmlauthor{Pauching Yap}{ucl}
\icmlauthor{Hippolyt Ritter}{ucl}
\icmlauthor{David Barber}{ucl,ati}
\end{icmlauthorlist}

\icmlaffiliation{ucl}{Department of Computer Science, University College London, London, United Kingdom}
\icmlaffiliation{ati}{Alan Turing Institute, London, United Kingdom}

\icmlcorrespondingauthor{Pauching Yap}{p.yap@cs.ucl.ac.uk}

\icmlkeywords{Bayesian Online Learning, few-shot learning, meta-learning, catastrophic forgetting}

\vskip 0.3in
]



\printAffiliationsAndNotice{}  

\begin{abstract}
Neural networks are known to suffer from catastrophic forgetting when trained on sequential datasets. While there have been numerous attempts to solve this problem in large-scale supervised classification, little has been done to overcome catastrophic forgetting in few-shot classification problems. We demonstrate that the popular gradient-based model-agnostic meta-learning algorithm (MAML) indeed suffers from catastrophic forgetting and introduce a Bayesian online meta-learning framework that tackles this problem. Our framework utilises Bayesian online learning and meta-learning along with Laplace approximation and variational inference to overcome catastrophic forgetting in few-shot classification problems. The experimental evaluations demonstrate that our framework can effectively achieve this goal in comparison with various baselines. As an additional utility, we also demonstrate empirically that our framework is capable of meta-learning on sequentially arriving few-shot tasks from a stationary task distribution.
\end{abstract}

\section{Introduction}
\label{sec:intro}

Few-shot classification \citep{Miller00, Li04, Lake11} focuses on learning to adapt to unseen classes (known as \newterm{novel classes}) with very few labelled examples from each class. Recent works show that meta-learning provides promising approaches to few-shot classification problems \citep{Santoro16, Finn17, Ravi17}. Meta-learning or learning-to-learn \citep{Schmidhuber87, Thrun98} takes the learning process a level deeper -- instead of learning from the labelled examples in the training classes (known as \newterm{base classes}), meta-learning learns the example-learning process. The training process in meta-learning that utilises the base classes is called the \newterm{meta-training} stage, and the evaluation process that reports the few-shot performance on the novel classes is known as the \newterm{meta-evaluation} stage.

\begin{figure}[ht]
\vskip 0.2in
\begin{center}
\centerline{\includegraphics[width=0.65\columnwidth]{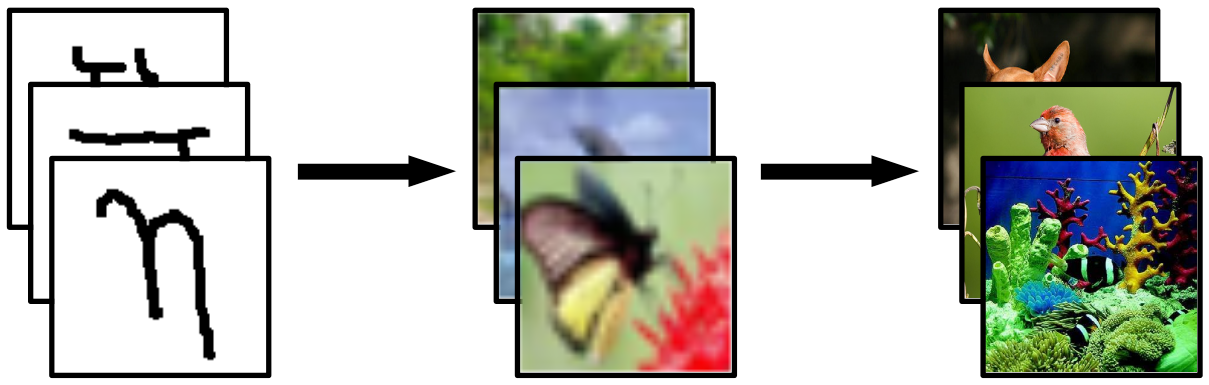}}
\caption{An example of the sequential few-shot classification problems with evident dataset distributional shift: \\ Omniglot $\rightarrow$ CIFAR-FS $\rightarrow$ \emph{mini}ImageNet.}
\label{fig:cf_eg}
\end{center}
\vskip -0.2in
\end{figure}

Despite being a promising solution to few-shot classification problems, meta-learning methods suffer from a limitation where a meta-learned model loses its few-shot classification ability on previous datasets as new ones arrive subsequently for meta-training. Some popular examples of the few-shot classification datasets are Omniglot \citep{Lake11}, CIFAR-FS \citep{Bertinetto19} and \emph{mini}ImageNet \citep{Vinyals16}. A meta-learned model is restricted to perform few-shot classification on a specific dataset, in the sense that the base and novel classes have to originate from the same dataset distribution. The current practice to few-shot classify the novel classes from different datasets is to meta-learn a model for each dataset separately \citep{Snell16, Vinyals16, Bertinetto19}. This paper considers meta-learning a single model for few-shot classification on multiple datasets with evident distributional shift that arrive sequentially for meta-training. \Figref{fig:cf_eg} gives an example of the sequential few-shot classification setting of concern.

We introduce a recursive framework to train a model that is applicable to a broader scope of few-shot classification datasets by overcoming catastrophic forgetting. Bayesian online learning (BOL) \citep{Opper98} provides a principled framework for the posterior of the model parameters, while model-agnostic meta-learning (MAML) \citep{Finn17} finds a good model parameter initialisation (called \newterm{meta-parameters}) that can quickly few-shot adapt to novel classes. Our framework incorporates BOL and meta-learning to give a recursive formula for the posterior of the meta-parameters as new few-shot datasets arrive. Taking a MAP estimate in implementation leads to Laplace approximation, whereas using a KL-divergence leads to variational inference. Our work builds on \citet{Ritter18} that combine BOL and Laplace approximation, and \citet{Nguyen18} that use variational inference with BOL to prevent forgetting in large-scale supervised classification.

\paragraph{Advantage of our framework:} An important reason to employ BOL over non-Bayesian approaches such as regret-based methods in an online setting is that BOL provides a grounded framework that suggests using the previous posterior as the prior recursively. BOL implicitly keeps a memory on previous knowledge via the posterior, in contrast to recent online meta-learning methods that explicitly accumulate previous data in a task buffer \citep{Finn19, Zhuang19}. Explicitly keeping a memory on previous data often triggers an important question: \emph{how should the carried-forward data be processed in future rounds, in order to accumulate knowledge?} \citet{Finn19} update the meta-parameters at each iteration using data sampled from the accumulated task buffer. This defeats the purpose of online learning, which by definition means to update the parameters each round using only the new data encountered.

\paragraph{Disadvantage of memorising past data:} Having to re-train on previous data to avoid forgetting also increases the training time as the data accumulate \citep{Finn19, He19}. Certainly one can clamp the amount of data at some maximal limit and sample from the buffer, but the final performance of such an algorithm would be dependent on the samples being informative and of good quality which may vary across different seed runs. In contrast to memorising the datasets, having an implicit memory via the posterior automatically deals with the question on how to process carried-forward data and allows a better knowledge accumulation process.

Below are the contributions we make in this paper:
\begin{itemize}
    \item We develop the Bayesian online meta-learning (\boml{}) framework for sequential few-shot classification problems. Under this framework we introduce the algorithms Bayesian online meta-learning with Laplace approximation (\bomla{}) and Bayesian online meta-learning with variational inference (\bomvi{}).
    \item We propose an approximation to the Fisher corresponding to \bomla{} that carries the desirable block-diagonal Kronecker-factored structure. 
    \item We demonstrate that \boml{} can overcome catastrophic forgetting in the sequential few-shot datasets setting with apparent distributional shift in the datasets.
    \item We demonstrate empirically that \boml{} can also continually learn to few-shot classify the novel classes in the sequential meta-training few-shot tasks setting.  
\end{itemize}

\section{Meta-Learning} \label{sec:metalearn}
Most meta-learning algorithms comprise an inner loop for example-learning and an outer loop that learns the example-learning process. Such algorithms often require sampling a meta-batch of tasks at each iteration, where a \newterm{task} from a stationary task distribution $p(\task)$ is formed by sampling a subset of classes from the pool of base classes or novel classes during meta-training or meta-evaluation respectively. The $N$-way $K$-shot task, for instance, refers to sampling $N$ classes and using $K$ examples per class for few-shot quick adaptation. 

An offline meta-learning algorithm learns a model only for a specific dataset $\mathfrak{D}$, which is divided into the set of base classes $\dtset$ and novel classes $\widehat{\dtset}$ for meta-training and meta-evaluation respectively. Upon completing meta-training on $\dtset$, the goal is to perform well on an unseen task $\widehat{\dtset}^{*}$ sampled from the novel set $\widehat{\dtset}$ after a quick adaptation on a small subset $\widehat{\dtset}^{*, S}$ (known as the \newterm{support set}) of $\widehat{\dtset}^{*}$. The performance of this unseen task is evaluated on the \newterm{query set} $\widehat{\dtset}^{*, Q}$, where $\widehat{\dtset}^{*, Q} = \widehat{\dtset}^{*} \backslash \widehat{\dtset}^{*, S}$. Since $\widehat{\dtset}$ is not accessible during meta-training, this support-query split is mimicked on the base set $\dtset$ for meta-training. 

\paragraph{Model-agnostic meta-learning:}
Each updating step of the well-known meta-learning algorithm MAML \citep{Finn17} aims to improve the ability of the meta-parameters to act as a good model initialisation for a quick adaptation on unseen tasks. Each iteration of the MAML algorithm samples $M$ tasks from the base class set $\dtset$ and runs a few steps of stochastic gradient descent (SGD) for an inner loop task-specific learning. The number of tasks sampled per iteration is known as the \newterm{meta-batch size}. For task $m$, the inner loop outputs the task-specific parameters $\tilde{\theta}^{m}$ from a $k$-step SGD quick adaptation on the objective $\mathcal{L}(\theta, \dtset^{m, S})$ with the support set $\dtset^{m, S}$ and initialised at $\theta$:
\begin{equation} \label{eq:maml_sgd_inner}
    \tilde{\theta}^{m} = SGD_{k}(\mathcal{L}(\theta, \dtset^{m, S})), 
\end{equation}
where $m = 1, \ldots, M$. The outer loop gathers all task-specific adaptations to update the meta-parameters $\theta$ using the loss $\mathcal{L}(\tilde{\theta}^{m}, \dtset^{m, Q})$ on the query set $\dtset^{m, Q}$. 

The overall MAML optimisation objective is
\begin{equation} \label{eq:maml_obj}
    \argmin_{\theta} \frac{1}{M} \sum_{m=1}^{M} \mathcal{L}(SGD_{k}(\mathcal{L}(\theta, \dtset^{m, S})), \dtset^{m, Q}).
\end{equation}

Like most offline meta-learning algorithms, MAML assumes a stationary task distribution during meta-training and meta-evaluation. Under this assumption, a meta-learned model is only applicable to a specific dataset distribution. When the model encounters a sequence of datasets with apparent distributional shift, it loses the few-shot classification ability on previous datasets as new ones arrive for meta-training. Our work aims to meta-learn a single model for few-shot classification on multiple datasets that arrive sequentially for meta-training. We achieve this goal by incorporating meta-learning into the BOL framework to give the \emph{Bayesian online meta-learning} (\boml{}) framework that considers the posterior of the meta-parameters.

\section{Bayesian Online Meta-Learning Framework Overview}
\label{sec:boml_overview}

Our central contribution is to extend the benefits of meta-learning to the BOL scenario, thereby training models that can generalise across tasks whilst dealing with parameter uncertainty in the setting of sequentially arriving datasets. 

\begin{figure}[ht]
\vskip 0.2in
\begin{center}
\centerline{\includegraphics[width=\columnwidth]{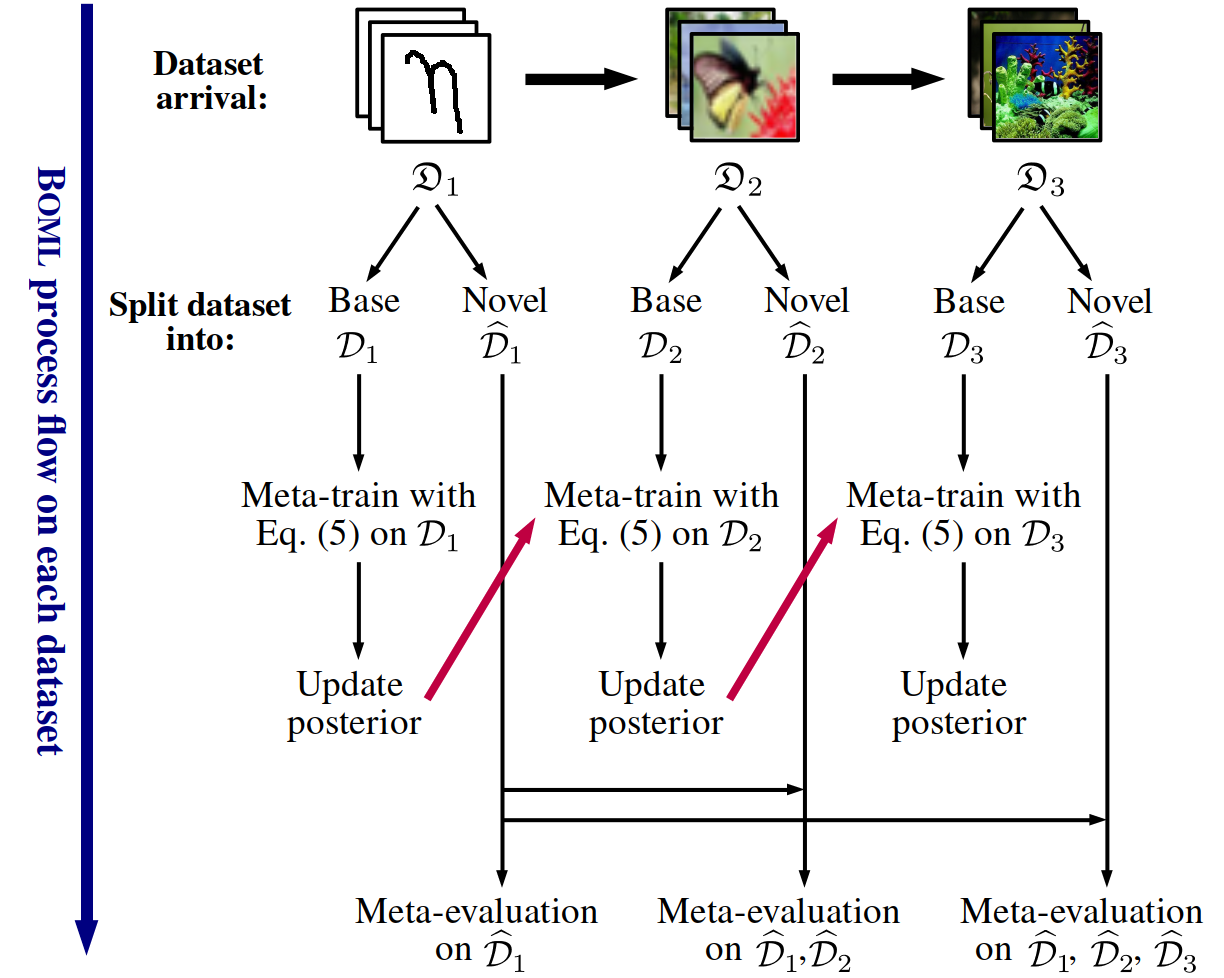}}
\caption{The \boml{} process flow for meta-training and meta-evaluation on an example sequence (Omniglot $\rightarrow$ CIFAR-FS $\rightarrow$ \emph{mini}ImageNet) when each dataset arrives. The arrows in \textcolor{purple}{purple} illustrate that the updated posterior is being brought forward for the next meta-training when a new dataset arrives.}
\label{fig:boml_process_flow}
\end{center}
\vskip -0.2in
\end{figure}

In this setting, meta-training occurs sequentially on the datasets $\mathfrak{D}_{1}, \ldots, \mathfrak{D}_{T}$. Each dataset $\mathfrak{D}_{i}$ can be seen as a knowledge domain with an associated underlying task distribution $p(\task_{i})$. A newly-arrived $\mathfrak{D}_{t+1}$ is separated into the base class set $\dtset_{t+1}$ and novel class set $\widehat{\dtset}_{t+1}$ for meta-training and meta-evaluation respectively, where the tasks in these two stages are drawn from the task distribution $p(\task_{t+1})$. Notationally, let $\dtset^{S}_{t+1}$ and $\dtset^{Q}_{t+1}$ denote the collection of support sets and query sets respectively from $\dtset_{t+1}$, so that $\dtset_{t+1} = \dtset^{S}_{t+1} \cup \dtset^{Q}_{t+1}$. Using Bayes' rule on the posterior gives the recursive formula
\begin{align}
    p & ( \theta | \dtset_{1:t+1}) \notag \\
        & \propto p(\dtset^{S}_{t+1}, \dtset^{Q}_{t+1} | \theta) \, p(\theta | \dtset_{1:t}) \label{eq:boml_l1} \\
        & = p(\dtset^{Q}_{t+1} | \theta, \dtset^{S}_{t+1}) \, p(\dtset^{S}_{t+1} | \theta) \, p(\theta | \dtset_{1:t}) \\
        & = \int p(\dtset^{Q}_{t+1} | \tilde{\theta}) \, p(\tilde{\theta} | \theta, \dtset^{S}_{t+1}) \, d \tilde{\theta} \, p(\dtset^{S}_{t+1} | \theta) \, p(\theta | \dtset_{1:t}) \label{eq:boml_metalearn}
\end{align}
where \Eqref{eq:boml_l1} follows from the assumption that each dataset is independent given $\theta$. \Figref{fig:boml_process_flow} illustrates the \boml{} process flow for meta-training and meta-evaluation as datasets arrive sequentially.

From the meta-learning perspective, the parameters $\tilde{\theta}$ introduced in \Eqref{eq:boml_metalearn} can be viewed as the task-specific parameters in MAML. There are various choices for the distribution $p(\tilde{\theta} | \theta, \dtset^{S}_{t+1})$ in \Eqref{eq:boml_metalearn}. In particular if we choose to set it as the deterministic function of taking several steps of SGD on loss $\mathcal{L}$ with the support set collection $\dtset^{S}_{t+1}$ and initialised at $\theta$, we have
\begin{equation} \label{eq:sgd_inner}
   p(\tilde{\theta} | \theta, \dtset^{S}_{t+1}) = \delta ( \tilde{\theta} - SGD_{k}(\mathcal{L}(\theta, \dtset^{S}_{t+1}) ) ),
\end{equation}
where $\delta(\cdot)$ is the Dirac delta function. This recovers the MAML inner loop with SGD quick adaptation in \Eqref{eq:maml_sgd_inner}. The recursion given by \Eqref{eq:boml_metalearn} forms the basis of our approach and the remainder of this paper explains how we implement this.

\section{Implementation} 
The posterior in \Eqref{eq:boml_metalearn} is typically intractable for modern neural network architectures. This leads to the requirement for a good approximate posterior. This section demonstrates how we arrive at the algorithms Bayesian online meta-learning with Laplace approximation (\bomla) and Bayesian online meta-learning with variational inference (\bomvi) by implementing Laplace approximation and variational continual learning (VCL) respectively to the \boml{} posterior in \Eqref{eq:boml_metalearn}. We give a mini tutorial in \Appref{sec:bg} on BOL, VCL and Laplace approximation.

\subsection{\bomla} \label{sec:bomla}
As described in \Appref{sec:lapl_approx}, Laplace approximation justifies the use of a Gaussian approximate posterior by Taylor expanding the log-posterior around a mode up to the second order. The second order term corresponds to the log-probability of a Gaussian distribution. The \boml{} framework in \Eqref{eq:boml_metalearn} with a Gaussian approximate posterior $q$ of mean and precision $\phi_t = \{ \mu_t, \Lambda_t \}$ from the Laplace approximation gives a MAP estimate:
\begin{equation} \label{eq:bomla_la_map_est}
    \theta^{*} = \argmax_{\theta} \big\{ 
    \log \Bar{p}_{\theta}
    + \log p(\dtset^{S}_{t+1} | \theta)
    + r_{\theta} \big\},
\end{equation}
where 
\begin{align*}
    & \Bar{p}_{\theta} = \int p(\dtset^{Q}_{t+1} | \tilde{\theta}) p(\tilde{\theta} | \theta, \dtset^{S}_{t+1}) \, d \tilde{\theta}, \\
    & r_{\theta} = - \frac{1}{2} (\theta - \mu_t)^{T} \Lambda_t (\theta - \mu_t).
\end{align*}
For an efficient optimisation, we use the deterministic $\tilde{\theta}$ in \Eqref{eq:sgd_inner} which leads to minimising the objective
\begin{equation} \label{eq:obj_bomla_seqdataset}
    f_{t+1}^{\bomla{}} (\theta, \mu_{t}, \Lambda_{t}) =  \Bar{f}_{\theta}^{(1)} + \Bar{f}_{\theta}^{(2)} - r_{\theta},
\end{equation}
where 
\begin{align*}
    & \Bar{f}_{\theta}^{(1)} = - \frac{1}{M} \sum_{m=1}^{M} \log p(\dtset^{m, Q}_{t+1} | \tilde{\theta}^{m}), \\ 
    & \Bar{f}_{\theta}^{(2)} = - \frac{1}{M} \sum_{m=1}^{M} \log p(\dtset^{m, S}_{t+1} | \theta),
\end{align*}
with $\tilde{\theta}^{m} = SGD_{k}(\mathcal{L}(\theta, \dtset^{m, S}_{t+1}))$ for $m = 1, \ldots, M$ and $M$ denotes the number of tasks sampled per iteration. The first term $\Bar{f}_{\theta}^{(1)}$ in \Eqref{eq:obj_bomla_seqdataset} corresponds to the MAML objective in \Eqref{eq:maml_obj} with a cross-entropy loss, the second term $\Bar{f}_{\theta}^{(2)}$ can be viewed as the pre-adaptation loss on the support set and the last term $r_{\theta}$ can be seen as a regulariser.

We discover that the Laplace approximation method provides a well-fitted meta-training framework for \boml{} in \Eqref{eq:boml_metalearn}. Each updating step in the approximation procedure can be modified to correspond to the meta-parameters for few-shot classification, instead of the model parameters for large-scale supervised classification.

\subsection{Hessian approximation} \label{sec:kfac_bomla}

We calculate a block-diagonal Kronecker-factored Hessian approximation in order to update the precision $\Lambda_{t}$, as explained in \Appref{sec:bd-kfac}. 

The Hessian matrix corresponding to the first term of the \bomla{} objective in \Eqref{eq:obj_bomla_seqdataset} is 
\begin{equation} \label{eq:hessian_la_maml}
    \widetilde{H}^{ij}_{t+1} = \frac{1}{M} \sum_{m=1}^{M} - \frac{\partial ^ 2}{\partial \theta^{(i)} \partial \theta^{(j)}} \log p(\dtset_{t+1}^{m, Q} | \tilde{\theta}^{m}) ) \biggr \rvert_{\theta = \mu_{t+1}}.
\end{equation}
It is worth noting that the \bomla{} Hessian deviates from the original Laplace approximation Hessian in \Appref{sec:lapl_approx}, and it is necessary to derive an adjusted approximation to the Hessian with some further assumptions. 

The Hessian for a single data point can be approximated using the Fisher information matrix $F$ to ensure its positive semi-definiteness \citep{Martens15}:
\begin{equation} \label{eq:fim_bol}
    F = \E_{x, y} \bigg[ \frac{d}{d \theta} \log p(y | x, \theta) \frac{d}{d \theta} \log p(y | x, \theta)^{T} \bigg].
\end{equation}
Each $(x, y)$ pair for the Fisher in \bomla{} is associated to a task $m$. The Fisher information matrix $\widetilde{F}$ corresponding to the \bomla{} Hessian in \Eqref{eq:hessian_la_maml} for a single data point is
\begin{equation} \label{eq:fim_bomla_jac}
\begin{split}
    \widetilde{F} =  \frac{1}{M} \sum_{m=1}^{M}  & \E_{x, y} \bigg[ \bigg( \frac{\partial \tilde{\theta}^{m}}{\partial \theta} \bigg) \frac{d}{d \tilde{\theta}^{m}} \log p(y | x, \tilde{\theta}^{m}) \\ 
    &\times \frac{d}{d \tilde{\theta}^{m}} \log p(y | x, \tilde{\theta}^{m})^{T} \bigg( \frac{\partial \tilde{\theta}^{m}}{\partial \theta} \bigg)^T \bigg].
\end{split}
\end{equation}
The additional Jacobian matrix $\frac{\partial \tilde{\theta}^{m}}{\partial \theta}$ breaks the Kronecker-factored structure described by \citet{Martens15} for the original Fisher in \Eqref{eq:fim_bol}. 

The results in \citet{Finn17} show that the first step of the quick adaptation in $\tilde{\theta}^{m}$ contributes the largest change to the meta-evaluation objective, and the remaining adaptation steps give a relatively small change to the objective. We can reasonably assume a one-step SGD quick adaptation $\tilde{\theta}^{m} = \theta - \nabla_{\theta} \mathcal{L}(\theta, \dtset_{t+1}^{m, S})$ to approximate the Fisher, although in other parts of the implementation we use a few-step SGD. By imposing this assumption, the $(i,j)$-th entry of the Jacobian term can be interpreted as 
\begin{equation} \label{eq:jac_onestepsgd}
    \bigg( \frac{\partial \tilde{\theta}^{m}}{\partial \theta} \bigg)^{ij} = I^{ij} - \frac{\partial ^ 2 (- \log p(\dtset_{t+1}^{m, S} | \theta) )  }{\partial \theta^{(i)} \partial \theta^{(j)}},
\end{equation}
where $I$ is the corresponding identity matrix and the objective $\mathcal{L}$ involved is the negative log-likelihood. The Hessian for a single data point in the second term of \Eqref{eq:jac_onestepsgd} can be approximated by $F$ in \Eqref{eq:fim_bol} via the usual block-diagonal Kronecker-factored approximation. Putting the Jacobian back into \Eqref{eq:fim_bomla_jac} and expanding the factors give terms that multiply two or more Kronecker products together. The detailed derivation of $\widetilde{F}$ is explained in \Appref{sec:kfac_approx}. We introduce the posterior-regularising hyperparameter $\lambda$ when updating the precision: $\Lambda_{t+1} = \lambda \widetilde{H}_{t+1} + \Lambda_{t}$ and the rationale for introducing $\lambda$ is explained in \Appref{sec:hyperparam_precision_update}. The pseudo-code of the \bomla{} algorithm can be found in \Appref{sec:algo_bomla_bomvi}.

\subsection{\bomvi}
\label{sec:bomvi}

The VCL framework \citep{Nguyen18} is directly applicable to \boml{}. This section demonstrates how we arrive at the \bomvi{} algorithm by implementing VCL to the posterior of the \boml{} framework in \Eqref{eq:boml_metalearn}.

As described in \Appref{sec:vcl}, VCL approximates the posterior by minimising the KL-divergence over some pre-determined approximate posterior family $\mathcal{Q}$. Fitting the \boml{} posterior in \Eqref{eq:boml_metalearn} into the VCL framework gives the approximate posterior:
\begin{equation} \label{eq:bomvi_vcl_post}
    q(\theta | \phi_{t+1}) 
        = \argmin_{q \in \mathcal{Q}} \KL \big( q(\theta | \phi) \big \Vert \Breve{q}_{\phi_t} \big),
\end{equation}
where $\Breve{q}_{\phi_t} = \int p(\dtset^{Q}_{t+1} | \tilde{\theta}) \, p(\tilde{\theta} | \theta, \dtset^{S}_{t+1}) d \tilde{\theta} \,
    p(\dtset^{S}_{t+1} | \theta) q(\theta | \phi_{t})$.
Similar to \bomla{}, we use the deterministic $\tilde{\theta}$ in \Eqref{eq:sgd_inner}. This leads to minimising the objective
\begin{equation} \label{eq:obj_bomvi_seqdataset}
    f_{t+1}^{\bomvi{}}(\phi, \phi_{t})  
    =  \Breve{f}_{\phi}^{(1)} 
    + \Breve{f}_{\phi}^{(2)}
    + \Breve{r}_{\phi},
\end{equation}
where 
\begin{align*}
    & \Breve{f}_{\phi}^{(1)} = - \frac{1}{M} \sum_{m=1}^{M} \E_{q(\theta | \phi)} \big[ \log p(\dtset^{m, Q}_{t+1} | \tilde{\theta}^{m}) \big], \\
    & \Breve{f}_{\phi}^{(2)} = - \frac{1}{M} \sum_{m=1}^{M} \E_{q(\theta | \phi)} \big[ \log p(\dtset^{m, S}_{t+1} | \theta) \big], \\
    & \Breve{r}_{\phi} = \KL(q(\theta | \phi) \Vert q(\theta | \phi_{t})),
\end{align*}
with $\tilde{\theta}^{m} = SGD_{k}(\mathcal{L}(\theta, \dtset^{m, S}_{t+1}))$ for $m = 1, \ldots, M$ and $M$ denotes the number of tasks sampled per iteration. We use a Gaussian mean-field 
$q(\theta | \phi_{t}) = \prod_{d=1}^{D} N(\mu_{t, d}, \sigma_{t, d}^2)$, where $\phi_t = \{ \mu_{t,d}, \sigma_{t,d} \}_{d=1}^{D}$, $D = \text{dim}(\theta)$ and the objective in \Eqref{eq:obj_bomvi_seqdataset} is minimised over $\phi$. The pseudo-code of the \bomvi{} algorithm can be found in \Appref{sec:algo_bomla_bomvi}.

The term $\Breve{f}_{\phi}^{(1)}$ in \Eqref{eq:obj_bomvi_seqdataset} is rather cumbersome to estimate in optimisation. To compute its Monte Carlo estimate, we have to generate samples $\theta_{r} \sim q$ for $r = 1, \ldots, R$, and run a quick adaptation on each sampled meta-parameters $\theta_{r}$ before evaluating their log-likelihoods. This is computationally intensive and gives an estimator with large variance. We propose a workaround by modifying the inner loop SGD quick adaptation, and the details can be found in \Appref{sec:bomvi_mc_est}.

\section{Related Work} \label{sec:related}

\subsection{Online Meta-Learning} 

\paragraph{Regret minimisation:} 
The goal of this setting is to minimise the regret function, and the assumptions are made on the loss function rather than the task distribution. Recent works \citet{Finn19} and \citet{Zhuang19} belong to this category, where the aim is to compete with the best meta-learner and supersede it. These methods accumulate data as they arrive and meta-learn using all data acquired so far. Data accumulation is not desirable as the algorithmic complexity of training grows with the amount of data accumulated, and training time increases as new data arrive \citep{Finn19, He19}. The agent will eventually run out of memory for a long sequence of data. The \boml{} framework on the other hand is advantageous, as it only takes the current data and the posterior of the meta-parameters into consideration during optimisation. This gives a framework with an algorithmic complexity independent of the length of the dataset sequence.

\paragraph{Same underlying task distribution:} Sequential tasks are assumed to originate from the same underlying task distribution $p(\task)$ in this setting. \citet{Denevi19} introduce the online-within-online (OWO) and online-within-batch (OWB) settings, where OWO encounters tasks and examples within tasks sequentially while OWB encounters tasks sequentially but examples within tasks are in batch. Our work in the sequential datasets setting is novel in overcoming few-shot catastrophic forgetting, where the goal is to few-shot classify unseen tasks drawn from a sequence of distributions $p(\task_1), \ldots, p(\task_T)$ as explained in \Secref{sec:boml_overview}. \citet{He19}, \citet{Harrison19} and \citet{Jerfel19} look into continual meta-learning for a non-stationary task distribution where the task boundaries are unknown to the model. \citet{Jerfel19} consider a latent task structure to adapt to the non-stationary task distribution.

\subsection{Offline Meta-Learning} 
Previous meta-learning works attempt to solve few-shot classification problems in an offline setting, under the assumption of having a stationary task distribution during meta-training and meta-evaluation. A single meta-learned model is aimed to few-shot classify one specific dataset with all base classes of the dataset readily available in a batch for meta-training. There are two general frameworks for the offline meta-learning setting:

\paragraph{Probabilistic:} The MAML algorithm can be cast into a probabilistic inference problem \citep{Finn18} or with a hierarchical Bayesian structure \citep{Grant18, Yoon18}. \citet{Yoon18} use Stein Variational Gradient Descent (SVGD) for task-specific learning. \citet{Gordon19} implement probabilistic inference by considering the posterior predictive distribution with amortised networks. \citet{Grant18} discuss the use of a Laplace approximation in the task-specific inner loop to improve MAML using the curvature information. Although at first sight our work seems similar to \citet{Grant18} due to the use of Laplace approximation, our work is clearly distinct in terms of goal and context. \citet{Grant18} use Laplace approximation at the task-specific level, whilst we use Laplace approximation at the meta-level for the meta-parameters approximate posterior. The formulation in \citet{Grant18} does not accumulate past experience, whereas our work enables few-shot learning on unseen tasks from multiple knowledge domains sequentially. 
    
\paragraph{Non-probabilistic:} Gradient-based meta-learning \citep{Finn17, Nichol18, Rusu19} updates the meta-parameters by accumulating the gradients of a meta-batch of task-specific inner loop updates. The meta-parameters will be used as a model initialisation for a quick adaptation on the novel tasks. Metric-based meta-learning \citep{Koch15, Vinyals16, Snell16} utilises the metric distance between labelled examples. This method assumes that base and novel classes are from the same dataset distribution, and the metric distance estimations can be generalised to the novel classes upon meta-learning the base classes.

\subsection{Continual Learning} 
Modern continual learning works \citep{Goodfellow13, Lee17, Zenke17} focus primarily on large-scale supervised learning, in contrast to our work that looks into continual few-shot classification across sequential datasets with evident distributional shift. \citet{Wen18} utilise few-shot learning to improve on overcoming catastrophic forgetting via logit matching on a small sample from the previous tasks. The online learning element in our work is closely related to recent works that overcome catastrophic forgetting in large-scale supervised classification \citep{Kirkpatrick17, Zenke17, Ritter18, Nguyen18}. In particular, our work builds on the online Laplace approximation method \citep{Ritter18}. Our work extends this to the meta-learning scenario to avoid forgetting in few-shot classification problems. \citet{Nguyen18} provide an alternative of using variational inference instead of Laplace approximation for approximating the posterior. Our work utilises this approach to adapt the variational method to approximate the posterior of the meta-parameters by adjusting the KL-divergence objective.

\section{Experiments} \label{sec:experiment}
\subsection{$N$-athlon} \label{sec:nathlon}
We implement \bomla{} and \bomvi{}\footnote{Implementation code is available at \url{https://github.com/pauchingyap/boml}} to the 5-way 1-shot \textbf{triathlon} and \textbf{pentathlon}
sequences. The experiment details and the datasets explanations are in \Appref{sec:exp_seqdataset}. We compare our algorithms to the following baselines:

\textbf{Train-On-Everything (TOE): } When a new dataset arrives for meta-training, we randomly re-initialise the meta-parameters and run MAML meta-training using \emph{all datasets encountered so far}. 

\textbf{Sequential MAML: } Upon the arrival of a new dataset, we run MAML to meta-train \emph{only on the newly-arrived dataset}.

\textbf{Follow The Meta-Leader (FTML): } We introduce a slight modification to FTML \citep{Finn19} on its evaluation method, as FTML is not designed for few-shot learning on unseen tasks. In our experiments, we apply \texttt{Update-Procedure} in FTML to the data from unseen tasks, rather than the data from the same training task as in the original FTML.

\subsubsection{Triathlon} \label{sec:triathlon}
\begin{figure}[ht]
\vskip 0.2in
\begin{center}
\centerline{\includegraphics[width=0.55\columnwidth]{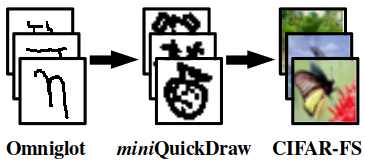}}
\caption{The triathlon 5-way 1-shot sequence in this experiment.}
\label{fig:triathlon}
\end{center}
\vskip -0.2in
\end{figure}

This experiment considers the few-shot triathlon sequence as in \Figref{fig:triathlon}.

\begin{figure}[ht]
\vskip 0.2in
\begin{center}
\centerline{\includegraphics[width=\columnwidth]{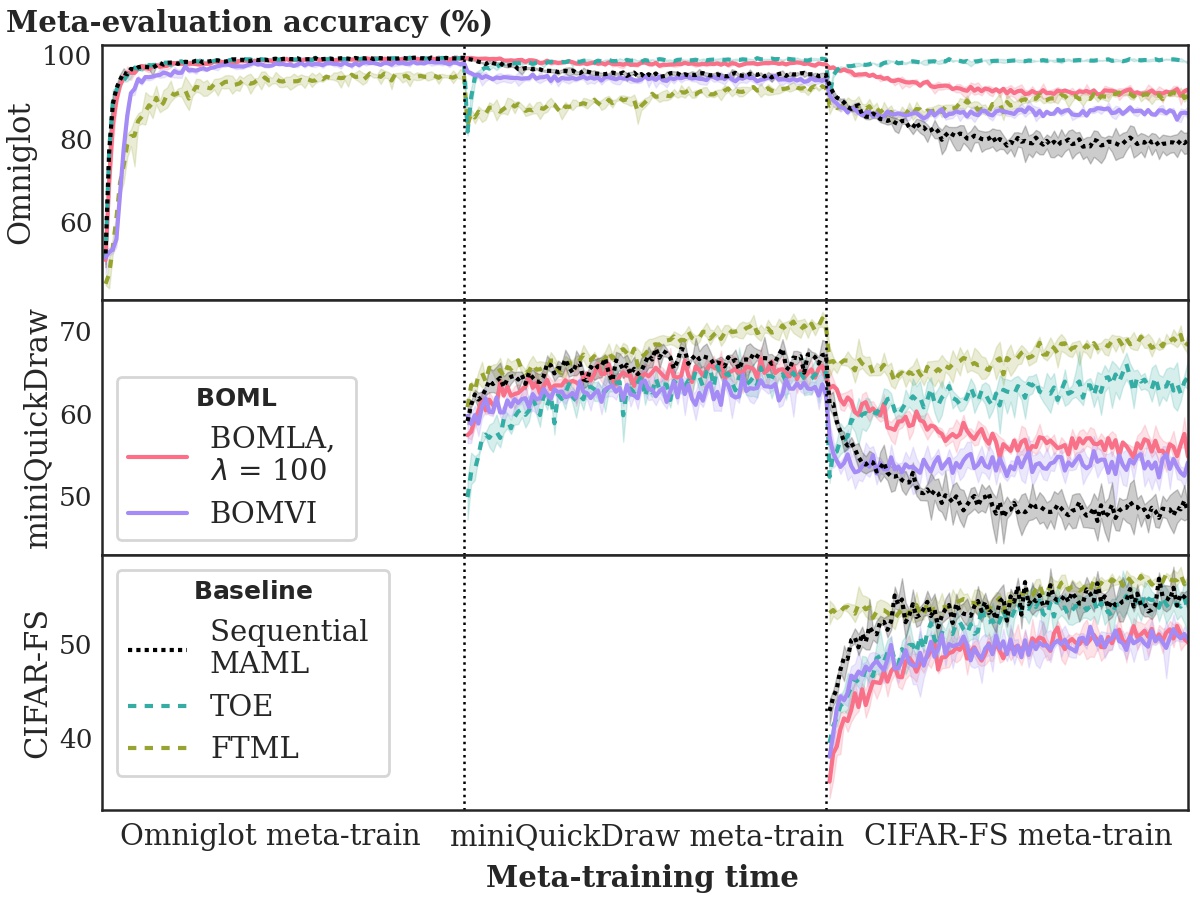}}
\caption{Meta-evaluation accuracy across 3 seed runs on each dataset along meta-training. Higher accuracy values in the off-diagonals indicate less forgetting.} 
\label{fig:ttl_result_icml2021}
\end{center}
\vskip -0.2in
\end{figure}

The distributional shift from Omniglot to \emph{mini}QuickDraw is less drastic, compared to the shift from \emph{mini}QuickDraw to CIFAR-FS. 
The result in \Figref{fig:ttl_result_icml2021} shows that \bomla{} and \bomvi{} are able to prevent catastrophic forgetting in both dataset transitions. \bomla{}, in particular, is able to proceed to the \emph{mini}QuickDraw meta-training phase with \textbf{almost no forgetting on Omniglot}. In other words, the meta-level pattern of Omniglot is retained throughout the meta-training period of \emph{mini}QuickDraw. There is a small trade-off in the performance of CIFAR-FS as \bomla{} and \bomvi{} avoid catastrophically forgetting Omniglot and \emph{mini}QuickDraw. Sequential MAML gives a noticeable drop in the performance of Omniglot and \emph{mini}QuickDraw when meta-training on CIFAR-FS. TOE is able to retain the few-shot performance as it has access to all previous datasets, whilst FTML gives a mixed performance. We elaborate on the result interpretation, the \bomla{}-\bomvi{} comparison and the choice of $\lambda$ along with the next experiment pentathlon, which resembles the setting of this experiment except with a more challenging dataset sequence.

\subsubsection{Pentathlon} \label{sec:pentathlon}
\begin{figure}[ht]
\vskip 0.2in
\begin{center}
\centerline{\includegraphics[width=\columnwidth]{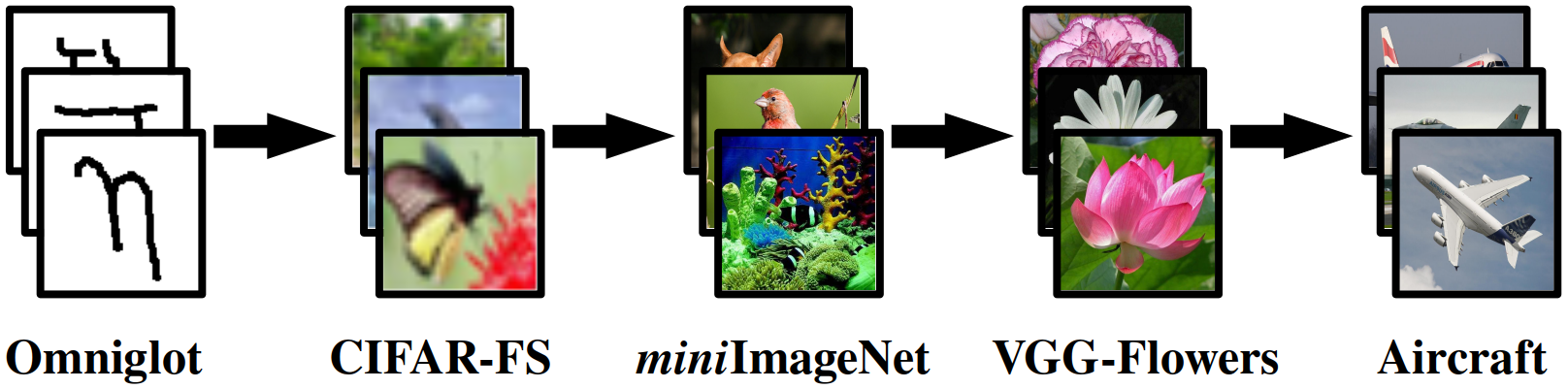}}
\caption{The pentathlon 5-way 1-shot sequence in this experiment.}
\label{fig:pentathlon}
\end{center}
\vskip -0.2in
\end{figure}

We implement \bomla{} and \bomvi{} to the more challenging pentathlon sequence as in \Figref{fig:pentathlon}.

\begin{figure*}[b!]
\vskip 0.2in
\begin{center}
\centerline{\includegraphics[width=\textwidth]{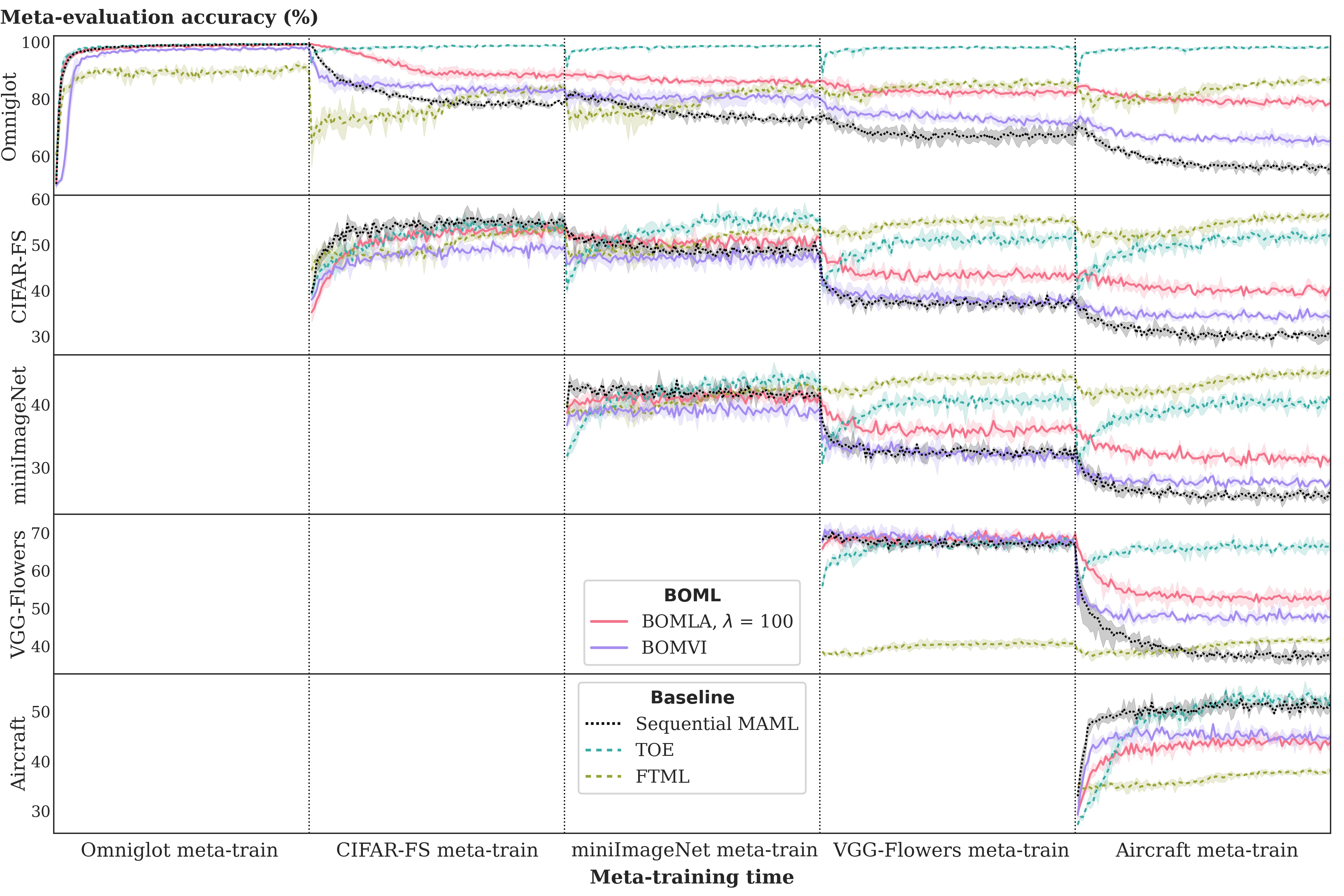}}
\caption{Meta-evaluation accuracy across 3 seed runs on each dataset along meta-training. Higher accuracy values indicate better results with less forgetting as we proceed to new datasets. \bomla{} with $\lambda = 100$ gives good performance in the off-diagonal plots (retains performances on previously learned datasets), and has a minor performance trade-off in the diagonal plots (learns less well on new datasets). Sequential MAML gives better performance in the diagonal plots (learns well on new datasets) but worse performance in the off-diagonal plots (forgets previously learned datasets). \bomvi{} is also able to retain performance on previous datasets, although it may be unable to perform as good as \bomla{} due to sampling and estimator variance.}
\label{fig:ptl_boml_result_icml2021}
\end{center}
\vskip -0.2in
\end{figure*}

\Figref{fig:ptl_boml_result_icml2021} shows that \bomla\ and \bomvi\ are able to prevent few-shot catastrophic forgetting in the pentathlon dataset sequence. TOE is also able to retain the few-shot performance as it has access to all datasets encountered so far. Since TOE learns all datasets from random re-initialisation each time it encounters a new dataset, the meta-training time required to achieve a similarly good meta-evaluation performance is longer compared to other runs. Sequential MAML catastrophically forgets the previously learned datasets but has the best performance on new datasets compared to other runs. FTML gives a mixed performance on different datasets. 

TOE and FTML can be memory-intensive as the dataset sequence becomes longer. They take the brute-force approach to prevent forgetting by memorising all datasets. Unlike TOE and FTML, our algorithms \bomla{} and \bomvi{} only take the newly-arrived dataset and the posterior of the meta-parameters into consideration during optimisation. This gives a framework with an algorithmic complexity independent of the length of the dataset sequence. 

\paragraph{\bomla{}-\bomvi{} comparison:}
As shown in \Figref{fig:ptl_boml_result_icml2021}, \bomla{} with appropriate $\lambda$ is superior to \bomvi{} in the performance. This is due to \bomla{} having a better posterior approximation than \bomvi{}. Whilst \bomla{} has a Gaussian approximate posterior with block-diagonal precision, \bomvi{} uses a Gaussian mean-field approximate posterior. \citet{Trippe17} compare the performances of variational inference with different covariance structures, and discover that variational inference with block-diagonal covariance performs worse than mean-field approximation. This is because the block-diagonal covariance in variational inference prohibits variance reduction methods such as local reparameterisation trick for Monte Carlo estimation. The variance of the Monte Carlo estimate has been proven problematic \citep{Kingma15lrt, Trippe17}. We address this issue in \Secref{sec:bomvi} and \Appref{sec:bomvi_mc_est} specifically to the meta-learning setting by modifying the inner loop quick adaptation. We analyse the change in the approximate posterior covariance in \Appref{sec:posterior_covariance}, as meta-training occurs sequentially on datasets from different knowledge domains.

\paragraph{Choosing $\lambda$:}
Tuning the posterior regulariser $\lambda$ mentioned in \Secref{sec:kfac_bomla} and \Appref{sec:hyperparam_precision_update} corresponds to balancing between a smaller performance trade-off on a new dataset and less forgetting on previous datasets. We compare \bomla{} with different $\lambda$ values and \bomvi{} in \Appref{sec:figs}. A larger $\lambda = 1000$ results in a more concentrated Gaussian posterior and is therefore unable to learn new datasets well, but can better retain the performances on previous datasets. A smaller $\lambda = 1$ on the other hand gives a widespread Gaussian posterior and learns better on new datasets by sacrificing the performance on the previous datasets. In both triathlon and pentathlon experiments, the value $\lambda = 100$ gives the best balance between old and new datasets. Ideally we seek for a good performance on both old and new datasets, but in reality there is a trade-off between retaining performance on old datasets and learning well on new datasets due to posterior approximation errors.

\subsection{Omniglot: Stationary Task Distribution} \label{sec:exp_omniglot}

\begin{figure}[ht]
\vskip 0.2in
\begin{center}
\centerline{\includegraphics[width=\columnwidth]{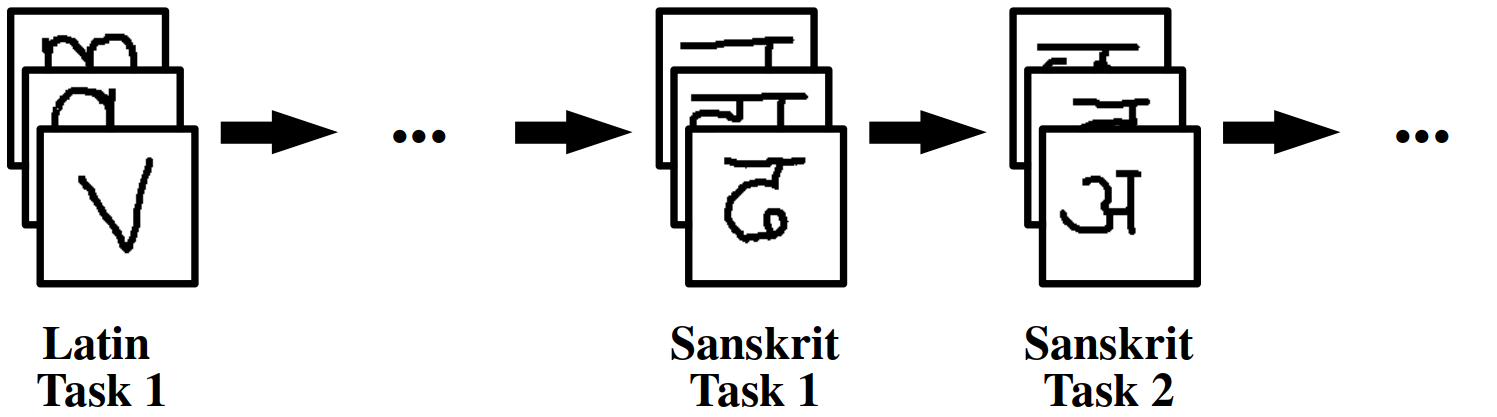}}
\caption{An example of the Omniglot task sequence for meta-training in this experiment.}
\label{fig:omni_seqtask}
\end{center}
\vskip -0.2in
\end{figure}

\begin{figure*}[h]
\begin{center}
\centerline{\includegraphics[width=\textwidth]{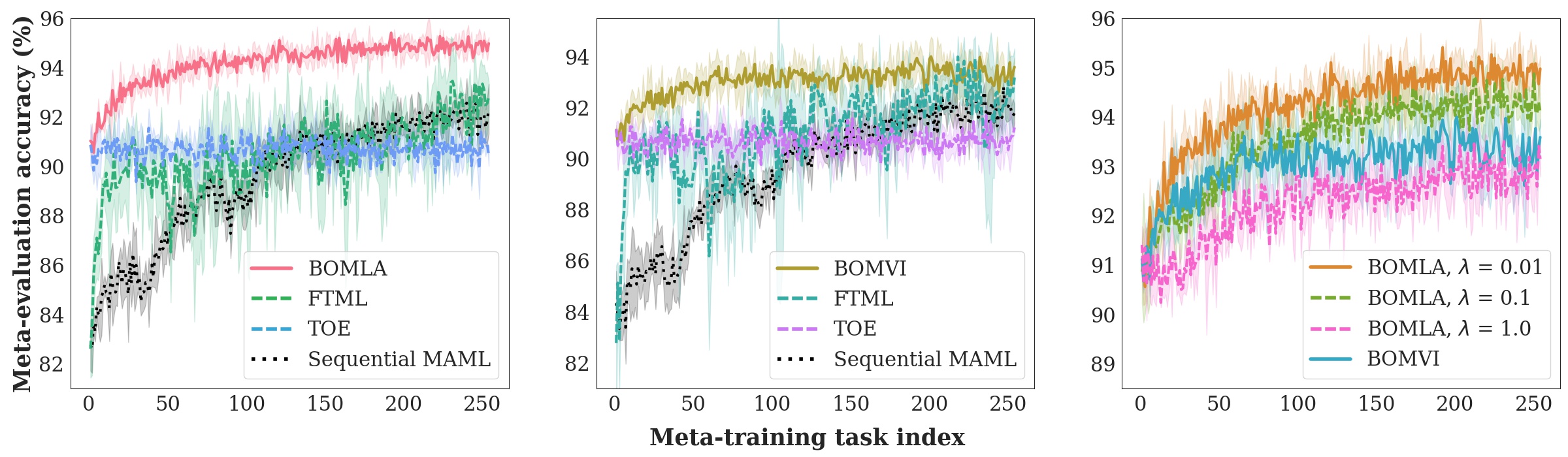}}
\caption{Meta-evaluation accuracy across 3 seed runs on the novel tasks along meta-training. \textbf{Left}: compares \bomla{} to the baselines, \textbf{centre}: compares \bomvi{} to the baselines, \textbf{right}: compares \bomla{} with different $\lambda$ values to \bomvi{}.}
\label{fig:seqtask_omni_icml2021}
\end{center}
\end{figure*}

In this experiment we demonstrate empirically that \boml{} can also continually learn to few-shot classify the novel classes in the sequential meta-training few-shot tasks setting, where all tasks originate from a stationary task distribution. This setting only involves \textbf{one dataset} $\mathfrak{D}$ with an associated underlying task distribution $p(\task)$, where $\mathfrak{D}$ is separated into the base and novel class sets. In this setting, $\dtset_{1}, \ldots, \dtset_{t+1}$ denote the non-overlapping tasks formed from the base class set and they arrive sequentially for meta-training. We show the corresponding modifications of \bomla{} and \bomvi{} under this setting in \Appref{sec:algo_bomla_bomvi} \Twoalgref{alg:bomla}{alg:bomvi}.

We run the sequential tasks experiment on the Omniglot dataset. To increase the difficulty level, we split the dataset based on the alphabets (super-classes) instead of the characters (classes) as in \Figref{fig:omni_seqtask}. The goal of this experiment is to classify the 5-way 5-shot novel tasks sampled from the meta-evaluation alphabets. The experimental details and the alphabet splits can be found in \Appref{sec:exp_seqtask}. We compare our algorithms to the baselines \textbf{TOE}, \textbf{Sequential MAML} and \textbf{FTML} similar to the $N$-athlon experiments but in the sequential tasks setting. 

As the tasks arrive sequentially for meta-training, \Figref{fig:seqtask_omni_icml2021} shows that \bomla{} and \bomvi{} can accumulate few-shot classification ability on the novel tasks over time. The knowledge acquired from previous meta-training tasks is carried forward in the form of a posterior, which is then used as a prior when a new task arrives for meta-training. Despite having access to all previous tasks, TOE shows no positive forward transfer in the meta-evaluation accuracy each time it encounters a new task. FTML and sequential MAML are inferior to \bomla{} and \bomvi{} in the performance. \bomla{} with $\lambda = 0.01$ gives the best performance in this experiment.

\subsection{Discussion on Baselines}

\citet{Finn19} discover that TOE does not explicitly learn the structure across tasks, thus unable to fully utilise the data. The TOE performance in our Omniglot experiment is coherent with the TOE result in \citet{Finn19}. The result figures in \citet{Finn19} show a TOE result similar to ours in the Omniglot experiment. In contrast, TOE in the $N$-athlon experiments performs well as it has access to drastically more data points than TOE in the Omniglot experiment, and samples numerous tasks from all previous datasets.

Sequential MAML in the $N$-athlon experiments suffers from catastrophic forgetting due to the apparent distributional shift in the datasets. The Omniglot experiment, on the other hand, has tasks originating from the same underlying distribution. As a result sequential MAML in this setting is able to accumulate few-shot ability, although it performs worse than \bomla{} and \bomvi{} as shown in \Figref{fig:seqtask_omni_icml2021} since there is only one task available at a time.

Since the original FTML is not aimed for unseen few-shot tasks and does not deal with sequential datasets setting as in the $N$-athlon experiments, we have to modify FTML as described in \Secref{sec:nathlon}. Sampling from previous tasks in the buffer is a key feature of the FTML algorithm. Certainly one can sample many tasks from the buffer to achieve perfect memory in the $N$-athlon experiments, but such a baseline setup has been taken into consideration by TOE. Therefore we choose to retain the online characteristic of the original FTML in our modified implementation.

\section{Conclusion}
We introduced the Bayesian online meta-learning (\boml{}) framework with two algorithms: \bomla{} and \bomvi{}. Our framework can overcome catastrophic forgetting in few-shot classification problems on datasets with evident distributional shift. \boml{} merged the BOL framework and meta-learning via Laplace approximation or variational continual learning. We proposed the necessary adjustments in the Hessian and Fisher approximation for \bomla{}, as we optimise the meta-parameters for few-shot classification instead of the usual model parameters in large-scale supervised classification. The experiments show that \bomla{} and \bomvi{} are able to retain the few-shot classification ability when trained on sequential datasets with apparent distributional shift, resulting in the ability to perform few-shot classification on multiple datasets with a single meta-learned model. \bomla{} and \bomvi{} are also able to continually learn to few-shot classify the novel tasks, as the meta-training tasks from a stationary distribution arrive sequentially for learning.

\section*{Acknowledgements}
We would like to thank the reviewers for their constructive comments, and Peter Hayes for the useful initial discussions.

\bibliography{reference}
\bibliographystyle{icml2021}

\newpage

\onecolumn

\appendix

\section{Background} \label{sec:bg}
This section provides a background explanation of using BOL to approximate the posterior of the model parameters and overcome catastrophic forgetting, commonly in large-scale supervised classification. We apply this approach to our recursion in \Eqref{eq:boml_metalearn}.

The posterior is typically intractable due to the enormous size of the modern neural network architectures. This leads to the requirement for a good approximation of the posterior of the model parameters. A particularly suitable candidate for this purpose is the Laplace approximation \citep{MacKay92, Ritter18iclr}, as it simply adds a quadratic regulariser to the training objective. Variational continual learning \citep{Nguyen18} is another possible method to obtain an approximation for the posterior of the model parameters.

\subsection{Bayesian Online Learning} \label{sec:bol}
Upon the arrival of the new dataset $ \mathfrak{D}_{t+1}$, we consider the posterior $p(\theta | \mathfrak{D}_{1:t+1})$ of the parameters $\theta$ of a neural network. Using Bayes' rule on the posterior gives the recursive formula
\begin{equation}
    p(\theta | \mathfrak{D}_{1:t+1}) \propto p(\mathfrak{D}_{t+1} | \theta) p(\theta | \mathfrak{D}_{1:t}) \label{eq:bol_l1}
\end{equation}
where \Eqref{eq:bol_l1} follows from the assumption that each dataset is independent given $\theta$. As the normalised posterior $p(\theta | \mathfrak{D}_{1:t})$ is usually intractable, it can be approximated by a parametric distribution $q$ with parameter $\phi_{t}$. The BOL framework consists of the \emph{update step} and the \emph{projection step} \citep{Opper98}. The update step uses the approximate posterior $q(\theta | \phi_{t})$ obtained from the previous step for an update in the form of \Eqref{eq:bol_l1}:
\begin{equation} \label{eq:bol_projetion}
    p(\theta | \mathfrak{D}_{1:t+1}, \phi_{t}) \propto p(\mathfrak{D}_{t+1} | \theta) q(\theta | \phi_{t}).
\end{equation}
The new posterior $p(\theta | \mathfrak{D}_{1:t+1}, \phi_{t})$ might not belong to the same parametric family as $q(\theta | \phi_{t})$. In this case, the new posterior has to be projected into the same parametric family to obtain $q(\theta | \phi_{t+1})$. \citet{Opper98} performs this projection by minimising the KL-divergence between the new posterior and the parametric $q$, while \citet{Ritter18} use Laplace approximation and \citet{Nguyen18} use variational inference. 

\subsection{Laplace Approximation} \label{sec:lapl_approx}

We consider finding a MAP estimate following from \Eqref{eq:bol_l1}:
\begin{equation} \label{eq:map_est}
    \theta^{*}_{t + 1}
    = \argmax_{\theta} \, p(\theta | \mathfrak{D}_{1:t+1}) 
    = \argmax_{\theta} \, \{  \log p(\mathfrak{D}_{t+1} | \theta) + \log p(\theta | \mathfrak{D}_{1:t}) \}.
\end{equation}
Since the posterior $p(\theta | \mathfrak{D}_{1:t})$ of a neural network is intractable except for small architectures, the unnormalised posterior $\Tilde{p}(\theta | \mathfrak{D}_{1:t})$ is considered instead. Performing Taylor expansion on the logarithm of the unnormalised posterior around a mode $\theta^{*}_{t}$ gives
\begin{equation} \label{eq:la_taylor}
    \log \Tilde{p}(\theta | \mathfrak{D}_{1:t}) \approx \log \Tilde{p}(\theta | \mathfrak{D}_{1:t}) \big \rvert_{\theta = \theta^{*}_{t}} - \frac{1}{2}(\theta - \theta^{*}_{t} )^{T} A_{t} (\theta - \theta^{*}_{t} ),
\end{equation}
where $A_t$ denotes the Hessian matrix of the negative log-posterior evaluated at $\theta^{*}_{t}$. The expansion in \Eqref{eq:la_taylor} suggests using a Gaussian approximate posterior. Given the parameter $\phi_t = \{ \mu_{t}, \Lambda_{t} \}$, a mean $\mu_{t+1}$ for step $t+1$ can be obtained by finding a mode of the approximate posterior as follows via a standard gradient-based optimisation:
\begin{equation} \label{eq:la_mode_mean}
    \mu_{t+1} = \argmax_{\theta} \, \bigg\{ \log p(\mathfrak{D}_{t+1} | \theta) - \frac{1}{2} (\theta - \mu_{t})^{T} \Lambda_{t} (\theta - \mu_{t}) \bigg\}.
\end{equation}
The precision matrix is updated as $\Lambda_{t+1} = H_{t+1} + \Lambda_{t}$, where $H_{t+1}$ is the Hessian matrix of the negative log-likelihood for $\mathfrak{D}_{t+1}$ evaluated at $\mu_{t+1}$ with entries
\begin{equation} \label{eq:hessian_la}
    H^{ij}_{t+1} = - \frac{\partial ^ 2}{\partial \theta^{(i)} \partial \theta^{(j)}} \log p( \mathfrak{D}_{t+1} | \theta) \biggr \rvert_{\theta = \mu_{t+1}}.
\end{equation}

For a neural network model, gradient-based optimisation methods such as SGD \citep{Robbins51} and Adam \citep{Kingma15} are the standard gradient-based methods in finding a mode for Laplace approximation in \Eqref{eq:la_mode_mean}. We show in Section~\ref{sec:bomla} that this provides a well-suited skeleton to implement Bayesian online meta-learning in \Eqref{eq:boml_metalearn} with the mode-seeking optimisation procedure.

\subsection{Block-Diagonal Hessian Approximation} \label{sec:bd-kfac}

Since the full Hessian matrix in \Eqref{eq:hessian_la} is intractable for large neural networks, we seek for an efficient and relatively close approximation to the Hessian matrix. Diagonal approximations \citep{Denker91, Kirkpatrick17} are memory and computationally efficient, but sacrifice approximation accuracy as they ignore the interaction between parameters. Consider instead separating the Hessian matrix into blocks where different blocks are associated to different layers of a neural network. A particular diagonal block corresponds to the Hessian for a particular layer of the neural network. The block-diagonal Kronecker-factored approximation \citep{Martens15, Grosse16, Botev17} utilises the fact that each diagonal block of the Hessian is Kronecker-factored for a single data point. This provides a better Hessian approximation as it takes the parameter interactions within a layer into consideration.

\subsubsection{Kronecker-Factored Approximation} \label{sec:kfac_approx}
Consider a neural network with $L$ layers and parameter $\theta = [\vectorise(W_{1})^{T}, \ldots, \vectorise(W_{L})^{T}]^{T}$ where $W_{\ell}$ is the weight of layer $\ell$ for $\ell = \{ 1, \ldots, L \}$ and $\vectorise$ denotes stacking the columns of a matrix into a vector. We denote the input of the neural network as $a_{0} = x$ and the output of the neural network as $h_{L}$. As the input passes through each layer of the neural network, we have the pre-activation for layer $\ell$ as $h_{\ell} = W_{\ell} a_{\ell - 1}$ and the activation as $a_{\ell} = f_{\ell}(h_{\ell})$ where $f_{\ell}$ is the activation function of layer $\ell$. If a bias vector is applicable in calculating the pre-activation of a layer, we append the bias vector to the last column of the weight matrix and append a scalar one to the last element of the activation. The gradient $g_{\ell}$ of loss $L_{\theta}(x, y) = - \log p(y | x, \theta)$ with respect to $h_{\ell}$ for an input-target pair $(x, y)$ is the pre-activation gradient for layer $\ell$.

\citet{Martens15} show that the $\ell$-th diagonal block $F_{\ell}$ of the Fisher information matrix $F$ can be approximated by the Kronecker product between the expectation of the outer product of the $(\ell - 1)$-th layer activation and the $\ell$-th layer pre-activation gradient:
\begin{align}
    F_{\ell} &= \E_{x, y}[ a_{\ell - 1} a_{\ell - 1}^{T} \otimes g_{\ell} g_{\ell}^{T} ] \\
        & \approx \E_{x}[a_{\ell - 1} a_{\ell - 1}^{T}] \otimes \E_{y|x}[g_{\ell} g_{\ell}^{T}] \\
        &= A_{\ell - 1} \otimes G_{\ell},
\end{align}

where $A_{\ell - 1} = \E_{x}[a_{\ell - 1} a_{\ell - 1}^{T}]$ and $G_{\ell} = \E_{y|x}[g_{\ell} g_{\ell}^{T}]$. \citet{Grosse16} extend the block-diagonal Kronecker-factored Fisher approximation for fully-connected layers to that for convolutional layers. For batch normalisation layers, we adopt the unit-wise approximation \citep{Osawa20}. The Gaussian log-probability term can be calculated efficiently without expanding the Kronecker product using the identity
\begin{equation}
    (A_{\ell - 1} \otimes G_{\ell}) \vectorise (W_{\ell} - W_{\ell}^{*}) = \vectorise (G_{\ell} (W_{\ell} - W_{\ell}^{*}) A_{\ell - 1}^{T}).
\end{equation}

As we mentioned in \Secref{sec:kfac_bomla}, approximating the Hessian with the one-step SGD inner loop assumption results in having terms that multiply two or more Kronecker products together. The $\ell$-th diagonal block of $\widetilde{F}$ in \Eqref{eq:fim_bomla_jac} is
\begin{equation} \label{eq:fim_bomla_jac_l-th_block}
    \widetilde{F}_{\ell} = \frac{1}{M} \sum_{m=1}^{M} ( I - A_{\ell - 1}^{m} \otimes G_{\ell}^{m} ) ( \widetilde{A}_{\ell - 1}^{m} \otimes \widetilde{G}_{\ell}^{m} ) ( I - A_{\ell - 1}^{m} \otimes G_{\ell}^{m} )^{T},
\end{equation}
where $\widetilde{A}_{\ell - 1}^{m} \otimes \widetilde{G}_{\ell}^{m}$ is the Kronecker product corresponding to the non-Jacobian terms in \Eqref{eq:fim_bomla_jac} for task $m$, and $A_{\ell - 1}^{m} \otimes G_{\ell}^{m}$ is the Kronecker product corresponding to the Hessian in \Eqref{eq:jac_onestepsgd}. We expand $\widetilde{F}_{\ell}$ using the Kronecker product property:
\begin{equation}
    ( A_{\ell - 1}^{m} \otimes G_{\ell}^{m})( \widetilde{A}_{\ell - 1}^{m} \otimes \widetilde{G}_{\ell}^{m} ) = A_{\ell - 1}^{m} \widetilde{A}_{\ell - 1}^{m} \otimes G_{\ell}^{m} \widetilde{G}_{\ell}^{m}.
\end{equation}
This gives 
\begin{equation}
    \widetilde{F}_{\ell} = \frac{1}{M} \sum_{m=1}^{M} \Big\{ \widetilde{A}_{\ell - 1}^{m} \otimes \widetilde{G}_{\ell}^{m} 
    - A_{\ell - 1}^{m} \widetilde{A}_{\ell - 1}^{m} \otimes G_{\ell}^{m} \widetilde{G}_{\ell}^{m} 
    - \widetilde{A}_{\ell - 1}^{m} (A_{\ell - 1}^{m})^T \otimes \widetilde{G}_{\ell}^{m} (G_{\ell}^{m})^T \\
    + A_{\ell - 1}^{m} \widetilde{A}_{\ell - 1}^{m} (A_{\ell - 1}^{m})^T \otimes G_{\ell}^{m} \widetilde{G}_{\ell}^{m} (G_{\ell}^{m})^T \Big\}.
\end{equation}
Finally, moving the meta-batch averaging into the Kronecker factors gives the approximation:
\begin{equation}
    \widetilde{F}_{\ell} \approx  \widetilde{\bm{A}}_{\ell - 1} \otimes \widetilde{\bm{G}}_{\ell}
    - \bm{A}_{\ell - 1} \widetilde{\bm{A}}_{\ell - 1} \otimes \bm{G}_{\ell} \widetilde{\bm{G}}_{\ell}
    - \widetilde{\bm{A}}_{\ell - 1} (\bm{A}_{\ell - 1})^T \otimes \widetilde{\bm{G}}_{\ell} (\bm{G}_{\ell})^T \\
    + \bm{A}_{\ell - 1} \widetilde{\bm{A}}_{\ell - 1} (\bm{A}_{\ell - 1})^T \otimes \bm{G}_{\ell} \widetilde{\bm{G}}_{\ell} (\bm{G}_{\ell})^T,
\end{equation}
where $\widetilde{\bm{A}}_{\ell - 1} = \frac{1}{M} \sum_m \widetilde{A}_{\ell - 1}^{m}$, 
$\widetilde{\bm{G}}_{\ell} = \frac{1}{M} \sum_m \widetilde{G}_{\ell}^{m}$, 
$\bm{A}_{\ell - 1} \widetilde{\bm{A}}_{\ell - 1} = \frac{1}{M} \sum_m A_{\ell - 1}^{m} \widetilde{A}_{\ell - 1}^{m}$, and so on.

\subsubsection{Posterior Regularising Hyperparameter for Precision Update} \label{sec:hyperparam_precision_update}
\citet{Ritter18} use a hyperparameter $\lambda$ as a multiplier to the Hessian when updating the precision:
\begin{equation}\label{eq:la_lambda}
    \Lambda_{t+1} = \lambda H_{t+1} + \Lambda_{t}.
\end{equation}
In the large-scale supervised classification setting, this hyperparameter has a regularising effect on the Gaussian posterior approximation for a balance between having a good performance on a new dataset and maintaining the performance on previous datasets \citep{Ritter18}. A large $\lambda$ results in a sharply peaked Gaussian posterior and is therefore unable to learn new datasets well, but can prevent forgetting previously learned datasets. A small $\lambda$ in contrast gives a dispersed Gaussian posterior and allows better performance on new datasets by sacrificing the performance on the previous datasets. 

\subsection{Variational Continual Learning} \label{sec:vcl}
The variational continual learning method \citep{Nguyen18} also provides a suitable meta-training framework for \boml{} in \Eqref{eq:boml_metalearn}. Consider approximating the posterior $q$ by minimising the KL-divergence between the parametric $q$ and the new posterior as in the projection step in \Eqref{eq:bol_projetion}, where $q$ belongs to some pre-determined approximate posterior family $\mathcal{Q}$ with parameters $\phi_{t}$:
\begin{align}
    q(\theta | \phi_{t+1}) 
        &= \argmin_{q \in \mathcal{Q}} \KL(q(\theta | \phi) \Vert p( \mathfrak{D}_{t+1} | \theta) q(\theta | \phi_{t})) \label{eq:vcl_l1} \\
        &= \argmin_{q \in \mathcal{Q}} \big\{ - \E_{q(\theta | \phi)}[\log p( \mathfrak{D}_{t+1} | \theta)] + \KL(q(\theta | \phi) \Vert q(\theta | \phi_{t})) \big\}. \label{eq:vcl_l2}
\end{align}

The optimisation in \Eqref{eq:vcl_l2} leads to the objective
\begin{equation} \label{eq:vcl_obj}
    \phi_{t+1} = \argmin_{\phi} \big\{ - \E_{q(\theta | \phi)}[\log p( \mathfrak{D}_{t+1} | \theta)] + \KL(q(\theta | \phi) \Vert q(\theta | \phi_{t})) \big\}.
\end{equation}
One can use a Gaussian mean-field approximate posterior $q(\theta | \phi_{t}) = \prod_{d=1}^{D} N(\mu_{t, d}, \sigma_{t, d}^2)$, where $\phi_t = \{ \mu_{t,d}, \sigma_{t,d} \}_{d=1}^{D}$ and $D = \text{dim}(\theta)$. The first term in \Eqref{eq:vcl_obj} can be estimated via simple Monte Carlo with local reparameterisation trick \citep{Kingma15lrt}, and the second KL-divergence term has a closed form for Gaussian distributions.

\newpage
\section{Algorithms}

\subsection{\bomla{} and \bomvi{}} \label{sec:algo_bomla_bomvi}

\Algref{alg:bomla} gives the pseudo-code of the \bomla{} algorithm, with the corresponding variation for the \Secref{sec:exp_omniglot} Omniglot sequential tasks setting in \emph{brackets}. The algorithm is formed of three main elements: meta-training on a specific base set or task (line~\ref{algo:line_maml_start} -- \ref{algo:line_maml_end}), updating the Gaussian mean (line~\ref{algo:line_upd_mean}) and updating the Gaussian precision (line~\ref{algo:line_upd_precision_start} -- \ref{algo:line_upd_precision_end}). For the precision update, we approximate the Hessian using block-diagonal Kronecker-factored approximation.

\begin{algorithm}[tb] 
   \caption{Bayesian online meta-learning with Laplace approximation (\bomla{})}
   \label{alg:bomla}
\begin{algorithmic}[1]
   \State \textbf{Require:} sequential base sets \emph{(or tasks)} $\dtset_{1}, \ldots, \dtset_{T}$, learning rate $\alpha$, posterior regulariser $\lambda$, number of meta-training iterations \emph{(or epochs)} $J$, meta-batch size \emph{(or number of mini-batches)} $M$
   \State \textbf{Initialise:} $\mu_0$, $\Lambda_0$, $\theta$
   
   \For{$t = 1$ \textbf{to} $T$}
        \For{$i = 1, \ldots, J$} 
        \label{algo:line_maml_start} 
        \Comment{meta-training on base set \emph{(or task)} $\dtset_{t}$}
        
            \For{$m = 1$ \textbf{to} $M$}
                \State Sample task \emph{(or split the batch)} $\dtset_{t}^{m} = \dtset_{t}^{m, S} \cup \dtset_{t}^{m, Q}$
                \State Inner update $\tilde{\theta}^{m} = SGD_{k}(\mathcal{L}(\theta, \dtset_{t}^{m, S}))$ 
                \label{algo:line_sgd_inner}
            \EndFor
                
            \State Evaluate loss $f_{t}^{\bomla{}}(\theta, \mu_{t-1}, \Lambda_{t-1})$ in \Eqref{eq:obj_bomla_seqdataset}
            \State Outer update $\theta \leftarrow \theta - \alpha \nabla_{\theta} f_{t}^{\bomla{}}(\theta, \mu_{t-1}, \Lambda_{t-1})$
        \EndFor \label{algo:line_maml_end}
        
        \State Update mean $\mu_{t} \leftarrow \theta$ \label{algo:line_upd_mean} \Comment{update posterior mean}
        
        \State For sequential datasets, sample a number of tasks for Hessian approximation \label{algo:line_upd_precision_start}
        \State Run inner update in line~\ref{algo:line_sgd_inner} for each sampled task \emph{(or for each batch)}
        \State Approximate $\widetilde{H}_{t}$ with block-diagonal Kronecker-factored approximation to $\widetilde{F}$ in \Eqref{eq:fim_bomla_jac}
        \State Update precision $\Lambda_t \leftarrow \lambda \widetilde{H}_{t} + \Lambda_{t-1}$ 
        \label{algo:line_upd_precision_end} \Comment{update posterior precision}
   \EndFor
\end{algorithmic}
\end{algorithm}

\Algref{alg:bomvi} gives the pseudo-code of the \bomvi{} algorithm, with the corresponding variation for the \Secref{sec:exp_omniglot} Omniglot sequential tasks setting in \emph{brackets}. The algorithm is formed of two main elements: meta-training on a specific base set or task (line~\ref{algo:bomvi_line_maml_start} -- \ref{algo:bomvi_line_maml_end}) and updating the parameters of the Gaussian mean-field approximate posterior (line~\ref{algo:bomvi_line_upd_param}). 

\begin{algorithm}[tb]
   \caption{Bayesian online meta-learning with variational inference (\bomvi{})}
   \label{alg:bomvi}
\begin{algorithmic}[1]
   \State \textbf{Require:} sequential base sets \emph{(or tasks)} $\dtset_{1}, \ldots, \dtset_{T}$, learning rate $\alpha$, number of meta-training iterations \emph{(or epochs)} $J$, meta-batch size \emph{(or number of mini-batches)} $M$
   \State \textbf{Initialise:} $\phi_0 = \{ \mu_0, \sigma_0 \}$
   
   \For{$t = 1$ \textbf{to} $T$}
        \For{$i = 1, 2, \ldots, J$} 
        \label{algo:bomvi_line_maml_start} \Comment{meta-training on base set \emph{(or task)} $\dtset_{t}$}
            \For{$m = 1$ \textbf{to} $M$}
                \State Sample task \emph{(or split the batch)} $\dtset_{t}^{m} = \dtset_{t}^{m, S} \cup \dtset_{t}^{m, Q}$
                \State Inner update $\tilde{\theta}^{m} = SGD_{k}(\mathcal{L}(\theta, \dtset_{t}^{m, S}))$ 
                \label{algo:bomvi_line_sgd_inner}
            \EndFor
                
            \State Evaluate loss $f_{t}^{\bomvi{}}(\phi, \phi_{t-1})$ in \Eqref{eq:obj_bomvi_seqdataset}
            \State Outer update $\mu \leftarrow \mu - \alpha \nabla_{\mu} f_{t}^{\bomvi{}}(\phi, \phi_{t-1})$, and $\sigma \leftarrow \sigma - \alpha \nabla_{\sigma} f_{t}^{\bomvi{}}(\phi, \phi_{t-1})$
        \EndFor 
        \label{algo:bomvi_line_maml_end}
        
        \State Update $\mu_{t} \leftarrow \mu$ and $\sigma_{t} \leftarrow \sigma$ 
        \label{algo:bomvi_line_upd_param} \Comment{update posterior parameters}
   \EndFor
\end{algorithmic}
\end{algorithm}

\newpage
\subsection{\bomvi{} Monte Carlo Estimator} \label{sec:bomvi_mc_est}

Recall that the \bomvi{} objective is:
\begin{equation*}
    f_{t+1}^{\bomvi{}}(\phi, \phi_{t}) = - \frac{1}{M} \sum_{m=1}^{M} \E_{q(\theta | \phi)} \big[ \log p(\dtset^{m, Q}_{t+1} | \tilde{\theta}^{m}) \big]
    - \frac{1}{M} \sum_{m=1}^{M} \E_{q(\theta | \phi)} \big[ \log p(\dtset^{m, S}_{t+1} | \theta) \big] 
    + \KL(q(\theta | \phi) \Vert q(\theta | \phi_{t})),
\end{equation*}
where $\tilde{\theta}^{m} = SGD_{k}(\mathcal{L}(\theta, \dtset^{m, S}_{t+1}))$ for $m = 1, \ldots, M$. The Monte Carlo estimator for the first term of the \bomvi{} objective is difficult to compute, as every sampled meta-parameters $\theta_r$ for $r = 1, \ldots, R$ has to undergo a few-shot quick adaptation prior to the log-likelihood evaluation. As a consequence the estimator is prone to a large variance. Moreover, every quickly-adapted sample $\theta_r$ contributes to the meta-learning gradients of the posterior mean and covariance, resulting in a high computational cost when taking the meta-gradients.

To solve these impediments, we introduce a slight modification to the SGD quick adaptation $\widetilde{\theta}^{m}$. Instead of taking the gradients with respect to the sampled meta-parameters, we consider the gradients with respect to the \emph{posterior mean}. A one-step SGD quick adaptation, for instance, becomes:
\begin{equation}
    \widetilde{\theta}^{m} = \theta - \alpha \nabla_{\mu_t} \mathcal{L}(\mu_t, \dtset^{m, S}_{t+1}).
\end{equation}
This gives $\widetilde{\theta}^m \sim N(\widetilde{\mu}_t, \text{diag}(\sigma_t^2))$ where 
\begin{equation}
    \widetilde{\mu}_t = \mu_t - \alpha \nabla_{\mu_t} \mathcal{L}(\mu_t, \dtset^{m, S}_{t+1}),
\end{equation}
since $\theta \sim N(\mu_t, \text{diag}(\sigma_t^2))$. A quick adaptation with more steps works in a similar fashion. With this modification, we can calculate the Monte Carlo estimate for the first term using the local reparameterisation trick as usual.

\newpage
\section{Experiments}
\subsection{Triathlon and Pentathlon} \label{sec:exp_seqdataset}
In these experiments, we use the model architecture proposed by \citet{Vinyals16} that takes 4 modules with 64 filters of size $3 \times 3$, followed by a batch normalisation, a ReLU activation and a $2 \times 2$ max-pooling. A fully-connected layer is appended to the final module before getting the class probabilities with softmax. \Twotabref{tab:ptl_hyperparam_same_val}{tab:ptl_hyperparam} are the hyperparameters used in these experiment.

\paragraph{Omniglot:} 

Omniglot \citep{Lake11} comprises 1623 characters from 50 alphabets and each character has 20 instances. We use 1100 characters for meta-training, 100 characters for validation and the remaining for meta-evaluation. New classes with rotations in the multiples of $90 \degree$ are formed after splitting the characters as mentioned. 

\paragraph{\emph{mini}QuickDraw:} 

QuickDraw \citep{Ha17} comprises 345 categories of drawings collected from the players in the game ``Quick, Draw!''. We generate \emph{mini}QuickDraw by randomly sampling 1000 instances in each class of QuickDraw.

\paragraph{CIFAR-FS:} 

CIFAR-FS \citep{Bertinetto19} has 100 classes of objects and each class comprises 600 images. We use the same split as \citet{Bertinetto19}: 64 classes for meta-training, 16 classes for validation and 20 classes for meta-evaluation.

\paragraph{\emph{mini}ImageNet:} 

\emph{mini}ImageNet \citep{Vinyals16} takes 100 classes and 600 instances in each class from the ImageNet dataset. We use the same split as \citet{Ravi17}: 64 classes for meta-training, 16 classes for validation and 20 classes for meta-evaluation.

\paragraph{VGG-Flowers:} 

VGG-Flowers \citep{Nilsback08} comprises 102 different types of flowers as the classes. We randomly split 66 classes for meta-training, 16 classes for validation and 20 classes for meta-evaluation.

\paragraph{Aircraft:} 

Aircraft \citep{Maji13} is a fine-grained dataset consisting of 100 aircraft models as the classes and each class has 100 images. We randomly split 64 classes for meta-training, 16 classes for validation and 20 classes for meta-evaluation.

\begin{table}[h]
\caption{Hyperparameters for the triathlon and pentathlon experiments (same value for all datasets)}
\label{tab:ptl_hyperparam_same_val}
\vskip 0.15in
\begin{center}
\begin{small}
\begin{tabular}{lll}
\multicolumn{1}{c}{\bf Hyperparameter}  &\multicolumn{1}{c}{\bf \bomla{}} &\multicolumn{1}{c}{\bf \bomvi{}}
\\ \hline \\
Posterior regulariser $\lambda$                    &(various values)       &- \\
Precision initialisation values                 &$10^{-4} \sim 10^{-2}$ &- \\
Number of tasks sampled for Hessian approx.     &$5000$ &- \\
Covariance initialisation values                &-                      &$\exp(-5)$ \\
Number of Monte Carlo samples                   &-                      &$20$ \\
Meta-batch size $M$                             &$32$                   &$32$ \\
Number of query samples per class               &$15$                   &$15$ \\
Number of iterations per dataset                &$5000$                 &$5000$ \\
Outer loop optimiser                            &Adam                   &Adam \\
Outer loop learning rate                        &$0.001$                &$0.001$ \\
Number of tasks sampled for meta-evaluation     &$100$                  &$100$ \\
\end{tabular}
\end{small}
\end{center}
\vskip -0.1in
\end{table}

\begin{table}[h]
\caption{Hyperparameters for the triathlon and pentathlon experiments (individual datasets)}
\label{tab:ptl_hyperparam}
\vskip 0.15in
\begin{center}
\begin{small}
\begin{tabular}{p{3.6cm}llp{1.8cm}lp{2.2cm}p{1.1cm}}
\multicolumn{1}{c}{\bf Hyperparameter}  &\multicolumn{1}{c}{\bf Omniglot}
&\multicolumn{1}{c}{\bf \emph{mini}QuickDraw}
&\multicolumn{1}{c}{\bf CIFAR-FS} &\multicolumn{1}{c}{\bf \emph{mini}ImageNet} &\multicolumn{1}{c}{\bf VGG-Flowers} &\multicolumn{1}{c}{\bf Aircraft}
\\ \hline \\
Number of inner SGD steps in meta-training ($k$)    &$1$    &$3$   &$5$    &$5$   &$5$    &$5$\\
Inner SGD learning rate ($\alpha$)                  &$0.4$  &$0.2$ &$0.1$                  &$0.1$                  &$0.1$                    &$0.1$ \\
Outer learning rate decay schedule (none for \bomvi{})                  &-      &$\times 0.1$ halfway &$\times 0.1$ halfway   &$\times 0.1$ halfway   &$\times 0.1$ every 1000 iterations  &$\times 0.1$ halfway \\
Number of inner SGD steps in meta-evaluation        &$3$    &$5$   &$10$                   &$10$                   &$10$                     &$10$ \\
\end{tabular}
\end{small}
\end{center}
\vskip -0.1in
\end{table}

\newpage
\subsection{Pentathlon: Analysing the Change in Approximate Posterior Covariance}
\label{sec:posterior_covariance}

We visualise the covariance of the meta-parameters approximate posterior from \bomvi{} to better understand how the uncertainty in the algorithm prevents catastrophic forgetting in few-shot classification problems. We follow the pentathlon sequence going from left to right of the figure: Omniglot $\rightarrow$ CIFAR-FS $\rightarrow$ \emph{mini}ImageNet $\rightarrow$ VGG-Flowers $\rightarrow$ Aircraft. The Gaussian mean-field approximate posterior becomes increasingly concentrated in general as it learns on more datasets. This is especially true for the earlier layers (Conv 1 and Conv 2), meaning that the posterior progressively becomes very confident on the meta-parameters of the raw-level filters. The covariance for the layer closest to the classifier (Conv 4) remains large in general, although there are some filters with decreasing covariance. As the convolutional layer gets closer to the classifying layer, a larger fine-tuning in the meta-parameters is needed  \cite{Ravi19} to cope with few-shot tasks from different knowledge domains. The approximate posterior covariance from \bomla{} is too large for visualisation as it is block-diagonal. The \bomla{} covariance for each convolutional layer has dimension $D \times D$ where $D$ is the number of parameters in a convolutional layer. In theory, the \bomla{} covariance should also follow the same pattern as the \bomvi{} covariance.

\begin{figure*}[ht]
\begin{center}
\centerline{\includegraphics[width=\textwidth]{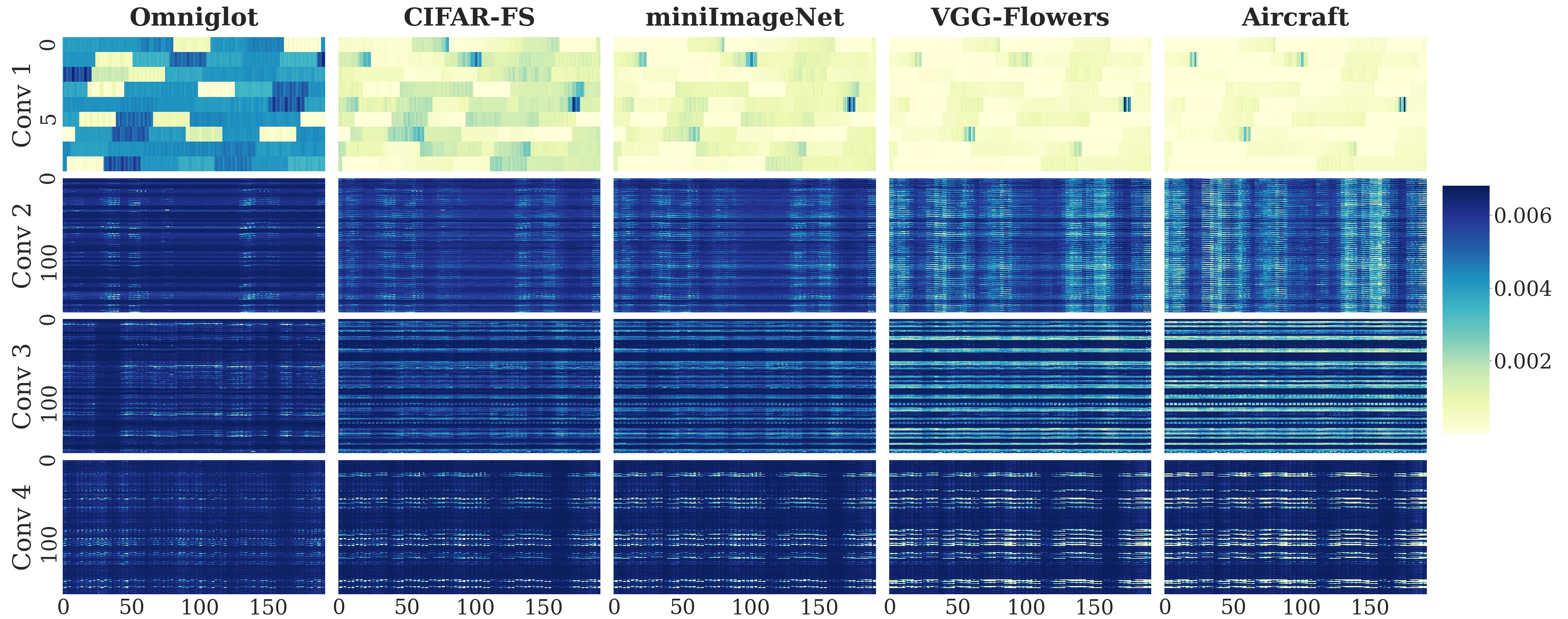}}
\caption{The change in the approximate posterior covariance after meta-training is completed on each dataset. Going from left to right are the pentathlon sequence of datasets. Going from top to bottom are the convolutional layers of the neural network which gets closer to the classifying layer.}
\end{center}
\label{fig:ptl_bomvi_cov_analysis_icml2021}
\end{figure*}

\newpage
\subsection{Pentathlon: the Comparison between \bomvi{} and \bomla{} with Different Values of $\lambda$} \label{sec:figs}

\begin{figure*}[ht]
\begin{center}
\centerline{\includegraphics[width=\textwidth]{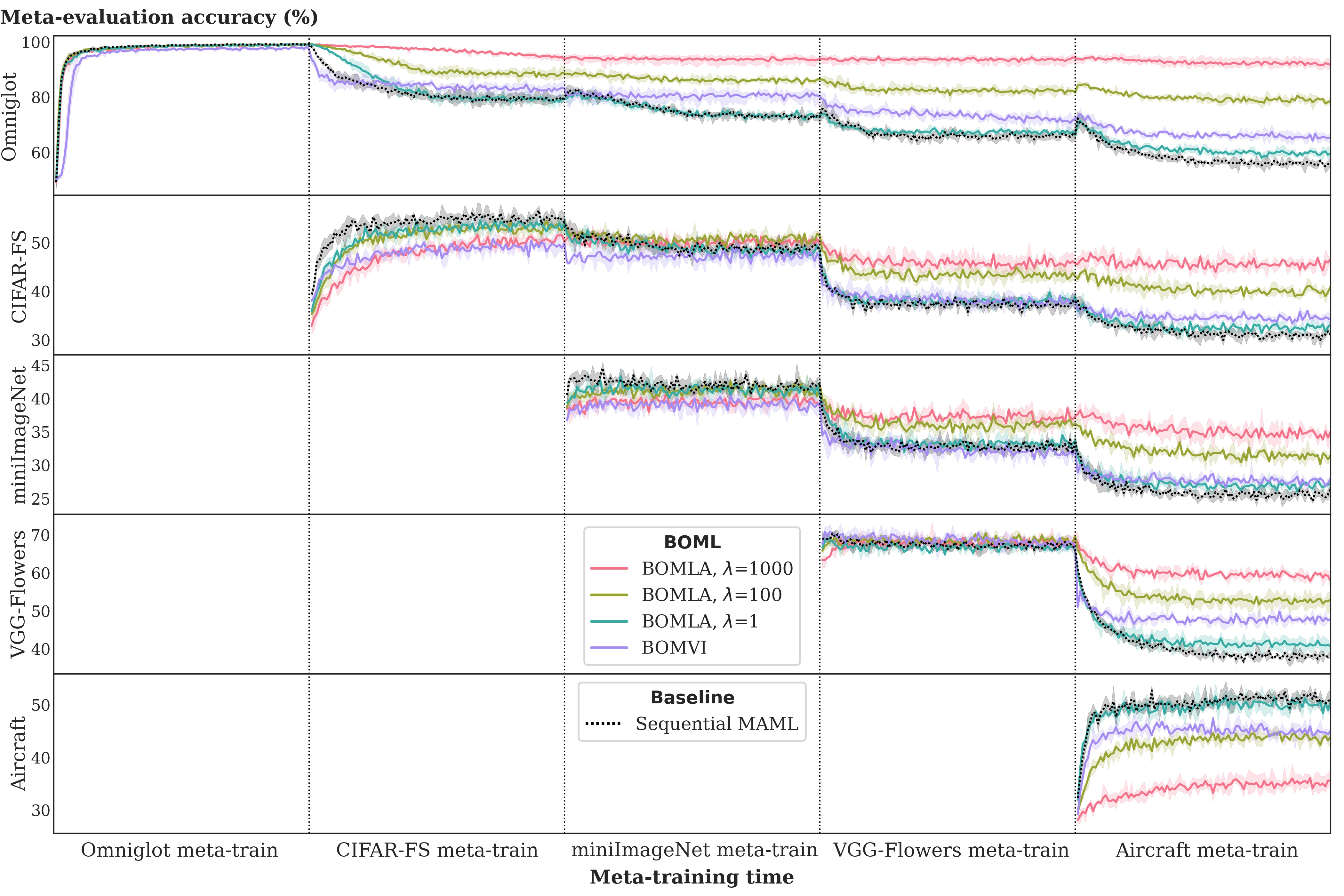}}
\caption{Meta-evaluation accuracy across 3 seed runs on each dataset along meta-training. Higher accuracy values indicate better results with less forgetting as we proceed to new datasets. \bomla{} with a large $\lambda = 1000$ gives better performance in the off-diagonal plots (retains performances on previously learned datasets) but worse performance in the diagonal plots (does not learn well on new datasets). A small $\lambda = 1$ gives better performance in the diagonal plots (learns well on new datasets) but worse performance in the off-diagonal plots (forgets previously learned datasets). \bomvi{} is also able to retain performance on previous datasets, although it may be unable to perform as good as \bomla{} due to sampling and estimator variance.}
\end{center}
\label{fig:ptl_lambda_result_icml2021}
\end{figure*}

Tuning the posterior regulariser $\lambda$ mentioned in \Secref{sec:kfac_bomla} and \Appref{sec:hyperparam_precision_update} corresponds to balancing between a smaller performance trade-off on a new dataset and less forgetting on previous datasets. As shown in the figure above, a larger $\lambda = 1000$ results in a more concentrated Gaussian posterior and is therefore unable to learn new datasets well, but can better retain the performances on previous datasets. A smaller $\lambda = 1$ on the other hand gives a widespread Gaussian posterior and learns better on new datasets by sacrificing the performance on the previous datasets. In this experiment, the value $\lambda = 100$ gives the best balance between old and new datasets. Ideally we seek for a good performance on both old and new datasets, but in reality there is a trade-off between retaining performance on old datasets and learning well on new datasets due to posterior approximation errors.

\newpage
\subsection{Omniglot: Sequential Tasks from a Stationary Task Distribution} \label{sec:exp_seqtask}
In this experiment, we use the model architecture proposed by \citet{Vinyals16} that takes 4 modules with 64 filters of size $3 \times 3$, followed by a batch normalisation, a ReLU activation and a $2 \times 2$ max-pooling. A fully-connected layer is appended to the final module before getting the class probabilities with softmax. \Tabref{tab:seqtask_omni_hyperparam} shows the hyperparameters used in this experiment.

The Omniglot dataset comprises 50 alphabets (super-classes). Each alphabet has numerous characters (classes) and each character has 20 instances. As the meta-training alphabets arrive sequentially, we form \textbf{non-overlapping} sequential tasks from each arriving alphabet, and the tasks also do not overlap in the characters. We use 35 alphabets for meta-training, 7 alphabets for validation and 8 alphabets for meta-evaluation. The alphabet splits are as follows:

35 alphabets for meta-training:
\begin{spverbatim}
Alphabet_of_the_Magi, Angelic, Armenian, Asomtavruli_(Georgian), Atlantean, Aurek-Besh, Avesta, Balinese, Bengali, Braille, Burmese_(Myanmar), Early_Aramaic, Grantha, Gujarati, Gurmukhi, Hebrew, Inuktitut_(Canadian_Aboriginal_Syllabics), Japanese_(hiragana), Japanese_(katakana), Kannada, Keble, Korean, Latin, Malayalam, Malay_(Jawi_-_Arabic), Manipuri, Mongolian, Ojibwe_(Canadian_Aboriginal_Syllabics), Old_Church_Slavonic_(Cyrillic), Oriya, Sanskrit, Sylheti, Tengwar, Tifinagh, ULOG
\end{spverbatim}

7 alphabets for validation:
\begin{spverbatim}
Anglo-Saxon_Futhorc, Arcadian, Blackfoot_(Canadian_Aboriginal_Syllabics), Cyrillic, Ge_ez, Glagolitic, N_Ko
\end{spverbatim}

8 alphabets for meta-evaluation:
\begin{spverbatim}
Atemayar_Qelisayer, Futurama, Greek, Mkhedruli_(Georgian), Syriac_(Estrangelo), Syriac_(Serto), Tagalog, Tibetan
\end{spverbatim}

\begin{table}[H]
\caption{Hyperparameters for the Omniglot sequential tasks experiment.}
\label{tab:seqtask_omni_hyperparam}
\vskip 0.15in
\begin{center}
\begin{small}
\begin{tabular}{lll}
\multicolumn{1}{c}{\bf Hyperparameter}  &\multicolumn{1}{c}{\bf \bomla{}} &\multicolumn{1}{c}{\bf \bomvi{}}
\\ \hline \\
Posterior regulariser $\lambda$                            &$0.01$                  &- \\
Precision initialisation values                         &$10^{-4} \sim 10^{-2}$ &- \\
Covariance initialisation values                        &-                      &$\exp(-10)$ \\
Number of Monte Carlo samples                           &-                      &$5$ \\
Number of mini-batches $M$                              &$1$                    &$1$ \\
Number of query samples per class (meta-evaluation)     &$15$                   &$15$ \\
Number of epochs per task                               &$50$                   &$50$ \\
Number of inner SGD steps in meta-training ($k$)        &$5$                    &$5$ \\
Inner SGD learning rate ($\alpha$)                      &$0.1$                  &$0.1$ \\
Outer loop optimiser                                    &Adam                   &Adam \\
Outer loop learning rate                                &$0.001$                &$0.001$ \\
Number of tasks sampled for meta-evaluation             &$100$                  &$100$ \\
Number of inner SGD steps in meta-evaluation ($k$)      &$10$                   &$10$ \\
\end{tabular}
\end{small}
\end{center}
\vskip -0.1in
\end{table}

\end{document}